%% file: main.tex
\documentclass[10pt,twocolumn,letterpaper]{article}
\input{setup/package}
\input{setup/macros}

\input{setup/symbols}

\input{setup/graphicspath}

\usepackage[accsupp]{axessibility}

\usepackage{iccv}      

\definecolor{iccvblue}{rgb}{0.21,0.49,0.74}

\def\paperID{5207} 
\def\confName{ICCV}
\def\confYear{2025}

\begin{document}

\title{ESSENTIAL: Episodic and Semantic Memory Integration \\ for Video Class-Incremental Learning}
\author{
Jongseo Lee$^{1}$\textsuperscript{*}  \quad
Kyungho Bae$^{2}$\textsuperscript{*} \quad
Kyle Min$^3$ \quad
Gyeong-Moon Park$^4$\textsuperscript{†} \quad
Jinwoo Choi$^1$\textsuperscript{†}\\
$^1$Kyung Hee University \quad
$^2$Danggeun Market Inc. \quad
$^3$Intel Labs \quad
$^4$Korea University \\
{\tt\small \{jong980812, kyungho.bae, jinwoochoi\}@khu.ac.kr, kyle.min@intel.com, gm-park@korea.ac.kr}
}

\twocolumn[{
\renewcommand\twocolumn[1][]{#1}
\maketitle
}]
\renewcommand{\thefootnote}{\fnsymbol{footnote}}
\footnotetext[1]{Equally contributed first authors.}
\footnotetext[2]{Corresponding authors.}
\input{0_abstract}
\input{1_introduction}
\input{2_prior}
\input{4_method}
\input{5_results}

\input{6_conclusion}

\paragraph{Acknowledgment.}
This work was supported in part by the Institute of Information \& Communications Technology Planning \& Evaluation (IITP) grant funded by the Korea Government (MSIT) under grant RS-2024-00353131 (20\%), RS-2021-0-02068 (Artificial Intelligence Innovation Hub, 20\%), and RS-2022-00155911 (Artificial Intelligence Convergence Innovation Human Resources Development (Kyung Hee University, 20\%)),
Additionally, it was supported by the Institute of Information \& Communications Technology Planning \& Evaluation(IITP)-ITRC(Information Technology Research Center) grant funded by the Korea government(MSIT)(IITP-2025-RS-2023-00259004, 20\%), and the National Research Foundation of Korea(NRF) grant funded by the Korea government(MSIT)(IRIS RS-2025-02216217, 20\%).
We thank Yuliang Zou for the valuable discussions and feedback.
%

\clearpage
\newpage

\maketitlesupplementary
\input{99_supp_content}

{\small
\bibliographystyle{ieee_fullname}
\bibliography{main}
}

\end{document}

%% file: setup/package.tex
\usepackage{subcaption}
\usepackage{float}
\usepackage[justification=justified]{caption}	
\usepackage{lscape}                                         
\usepackage{wrapfig}
\usepackage{bbding}
\usepackage{pifont}

\usepackage[lined,ruled,linesnumbered]{algorithm2e}

\usepackage{booktabs}                   
\usepackage{multirow}
\usepackage{makecell}
\usepackage{paralist}

\usepackage{bm}                          
\usepackage{epsfig}                      
\usepackage{times}
\usepackage{mathptmx}
\usepackage{mathtools}
\usepackage{amssymb}
\usepackage{amsmath}   

\usepackage{units}
\usepackage{xcolor}

\usepackage{comment}

\usepackage{url}  
\usepackage[pagebackref,breaklinks,colorlinks,allcolors=iccvblue]{hyperref}

\usepackage[capitalize]{cleveref}
\crefname{section}{Sec.}{Secs.}
\Crefname{section}{Section}{Sections}
\Crefname{table}{Table}{Tables}
\crefname{table}{Tab.}{Tabs.}

\usepackage{xspace}
\usepackage{setspace}

\usepackage{array}
\usepackage{colortbl}
\usepackage{arydshln}


%% file: setup/macros.tex
\def\eg{e.g.,~}               
\def\ie{i.e.,~}               
\def\vs{vs.~}                 

\newlength\paramarginsize
\newlength\figmarginsize
\newlength\tabmarginsize
\newlength\secmarginsize
\newlength\figcapmarginsize
\newlength\tabcapmarginsize

\newcommand{\paramargin}{\vspace{\paramarginsize}}

\newcommand{\figcapmargin}{\vspace{\figcapmarginsize}}

\newcommand{\green}{\textcolor{green}}
\definecolor{C1}{rgb}{0.34, 0.52, 0.86} 
\definecolor{C2}{rgb}{0.38, 0.87, 0.60} 
\definecolor{C3}{rgb}{0.86, 0.38, 0.35} 
\definecolor{C4}{RGB}{222,180,97}
\definecolor{before}{RGB}{176,146,238}
\definecolor{after}{RGB}{246,206,126}
\newcommand{\densec}[1]{\textcolor{C1}{#1}}
\newcommand{\recon}[1]{\textcolor{C2}{#1}}
\newcommand{\sparsec}[1]{\textcolor{C3}{#1}}
\newcommand{\random}[1]{\textcolor{C4}{#1}}
\newcommand{\mpage}[2]
{
\begin{minipage}{#1\linewidth}\centering
#2
\end{minipage}
}

\newcommand{\figcaption}[2]
{
\caption{
\textbf{#1.}  
#2            
}
}

\newcommand{\secref}[1]{Section~\ref{sec:#1}}
\newcommand{\figref}[1]{Figure~\ref{fig:#1}} 
\newcommand{\tabref}[1]{Table~\ref{tab:#1}}
\newcommand{\eqnref}[1]{\eqref{eq:#1}}

\long\def\ignorethis#1{}

\newcommand{\tb}[1]{\textbf{#1}}

\newbox\jsavebox%

\newcommand{\best}[1]{{\textbf{#1}}}
\newcommand{\second}[1]{{\underline{#1}}}

\def\ours{\texttt{ESSENTIAL}}
\def\cls{\texttt{CLS}}

%% file: setup/symbols.tex
\def\xi{\mathbf{x}_i}

\def\x{\mathbf{x}}
\def\X{\mathbf{X}}
\def\y{\mathbf{y}}

\def\S{\mathbf{S}}

\def\E{\mathbf{E}}
\def\z{\mathbf{z}}

\def\P{\mathbf{P}}

\def\SM{\mathbb{M}}
\def\episodic{\mathbb{M}^{E}}
\def\semantic{\mathbb{M}^{S}}
\def\SD{\mathbb{D}}

\def\dense{{\mathbf{S}}_\text{dense}}
\def\retrieved{{\tilde{\mathbf{S}}}_\text{dense}}
\def\sparse{{\mathbf{S}}_\text{sparse}}

\newcommand{\ffn}{\texttt{FFN}}
\newcommand{\mhca}{\texttt{MHCA}}
\newcommand{\softmax}{\texttt{Softmax}}

\newcommand{\lnorm}{\texttt{LN}}

\newcommand{\prompt}{\mathbf{P}}

%% file: setup/graphicspath.tex
\graphicspath{{figure}, {images}, {example}}

%% file: 0_abstract.tex
\begin{abstract}





In this work, we tackle the problem of video class-incremental learning (VCIL).
Many existing VCIL methods mitigate catastrophic forgetting by rehearsal training with a few temporally dense samples stored in episodic memory, which is memory-inefficient.
Alternatively, some methods store temporally sparse samples, sacrificing essential temporal information and thereby resulting in inferior performance.
To address this trade-off between memory-efficiency and performance, we propose EpiSodic and SEmaNTIc memory integrAtion for video class-incremental Learning (\ours{}).
%
%
\ours{} consists of episodic memory for storing temporally sparse features and semantic memory for storing general knowledge represented by learnable prompts.
We introduce a novel memory retrieval (MR) module that integrates episodic memory and semantic prompts through cross-attention, enabling the retrieval of temporally dense features from temporally sparse features.
We rigorously validate \ours{} on diverse datasets: UCF-101, HMDB51, and Something-Something-V2 from the TCD benchmark and UCF-101, ActivityNet, and Kinetics-400 from the vCLIMB benchmark.
Remarkably, with significantly reduced memory, \ours{} achieves favorable performance on the benchmarks. 

\end{abstract}

%% file: 1_introduction.tex
\section{Introduction}
\label{sec:intro}
\vspace{-0.5em}

With advances in deep learning, action recognition models trained on large annotated datasets have achieved remarkable performance~\cite{i3d,lin2019tsm,feichtenhofer2019slowfast,timesformer,arnab2021vivit,wu2022memvit,tong2022videomae,yang2023aim,pan2022st,lee2024cast}.
However, in more realistic scenarios such as class-incremental learning (CIL), action recognition models face significant challenges.
In the CIL setting, data for new action classes of interest is incrementally introduced, while data for previously learned classes is not available.
Deep learning models tend to overfit to the currently available data, leading to performance degradation on previously learned classes, a phenomenon known as catastrophic forgetting~\cite{mccloskey1989catastrophic,goodfellow2013empirical}.



\input{figure/fig_tradeoff}

We can mitigate catastrophic forgetting by rehearsal training with a few video samples or feature vectors stored in episodic memory.
As shown in \figref{motivation} (a), many prior methods store \emph{temporally dense} samples, \eg 16 frames per sample, in episodic memory to preserve temporal information of previous tasks~\cite{tcd,villa2022vclimb,villa2023pivot,stprompt}.
However, storing such temporally dense samples is memory-inefficient.
Memory-efficiency is crucial in real-world settings as we should consider the scalability, privacy, and legal concerns, as well as the performance of the methods.
A potential solution to improve memory-efficiency is to store \emph{temporally sparse} samples, \eg 1 or 2 frames per sample, in episodic memory~\cite{alssum2023just,framemaker}.
However, as shown in \figref{motivation} (b), this approach can lead to performance degradation as temporally sparse samples may lack temporal details essential for recognizing actions.
We desire a VCIL method that optimally balances high temporal resolution with memory-efficiency, achieving a better trade-off between memory-efficiency and performance.


To achieve a better trade-off between memory-efficiency and performance, it is worth considering how humans achieve efficient memory retrieval.
Rather than relying on a single memory system, the human brain integrates multiple complementary forms of memory (e.g., episodic and semantic) that interact to enable robust and efficient recall~\cite{wiggs1998neural,menon2002relating,eichenbaum2001hippocampus}.
%
%
This highlights that supporting episodic memory with an additional memory component is crucial for achieving a better trade-off between memory-efficiency and performance in VCIL.

\input{figure/fig_motivation}

Inspired by 
this,
we propose a novel method, \textbf{E}pi\textbf{S}odic and \textbf{SE}ma\textbf{NTI}c memory integr\textbf{A}tion for video class-incremental \textbf{L}earning (\ours{}) designed to achieve a better trade-off between memory-efficiency and performance.
To reduce memory consumption, \ours{} consists of two efficient memory components: 
i) episodic memory, which stores temporally \emph{sparse} features, 
and ii) semantic memory\footnote{
Used as a methodological term in this work, not strictly equivalent to semantic memory in cognitive neuroscience.
}, which stores general and abstract knowledge represented by lightweight learnable vectors, 
referred to as semantic prompts in our method.
%
Notably, we deliberately share a semantic prompt across multiple samples to promote the learning of general knowledge.
To effectively mitigate forgetting and achieve high performance, we introduce a novel memory retrieval (MR) module.
During \emph{rehearsal stage}, the MR module retrieves temporally dense features by integrating temporally sparse features from episodic memory with the lightweight prompt from semantic memory through cross-attention.
%
To enable the retrieval capabilities of the MR module, we employ static and temporal matching loss functions during each \emph{incremental training stage}.
Employing the matching loss functions encourages the retrieved temporally dense features to closely match the original temporally dense features.
As illustrated in \figref{motivation} (c), the distance between the retrieved features and the original temporally dense features is significantly smaller than the distance between the temporally sparse features and the original temporally dense features.
By effectively retrieving temporal information from temporally sparse features, the MR module enables a high memory-efficiency-performance trade-off as demonstrated in \figref{tradeoff}.
To validate the effectiveness of \ours{}, we carefully design a set of controlled experiments.
We evaluate each component of the proposed method on the temporal-biased Something-Something-V2 (SSV2) dataset~\cite{ssv2}.
We compare the proposed method against existing VCIL methods using diverse standard benchmarks: UCF-101~\cite{soomro2012ucf101}, ActivityNet~\cite{caba2015activitynet}, and Kinetics400~\cite{kinetics} from the vCLIMB~\cite{villa2022vclimb} benchmark~\cite{villa2022vclimb}, as well as UCF-101~\cite{soomro2012ucf101}, HMDB51~\cite{hmdb}, and SSV2~\cite{ssv2} from the TCD benchmark~\cite{villa2022vclimb}.
To facilitate future research, we plan to release our code.
To summarize, our major contributions are as follows:
\begin{itemize}
    \item We propose \ours{}, a \emph{memory-efficient} video class-incremental learning method that incorporates episodic memory for storing temporally sparse features and semantic memory for storing general knowledge.

    
    \item To achieve \emph{high performance}, we introduce a memory retrieval (MR) module that integrates episodic and semantic memory through cross-attention, enabling the \emph{retrieval of temporally dense features} by leveraging both temporally sparse features and general knowledge stored in semantic memory.
    
\end{itemize}

%% file: figure/fig_tradeoff.tex
\begin{figure}[t]
    \centering\centering
    \includegraphics[width=0.95\linewidth]{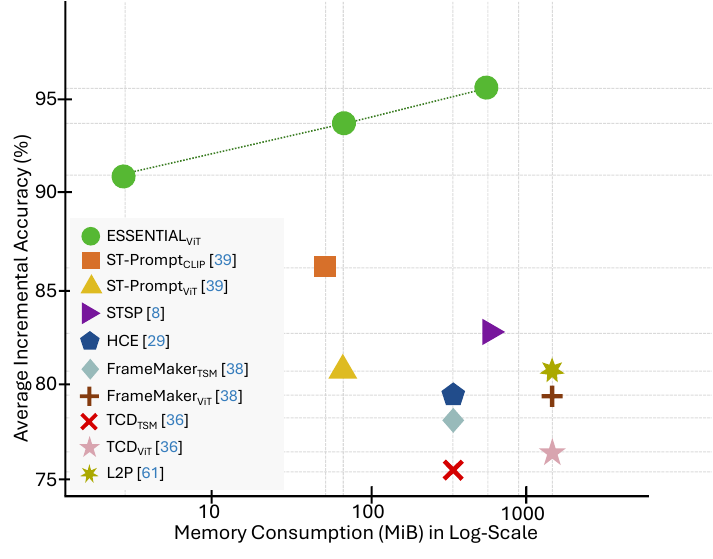}
\captionsetup{justification=justified, singlelinecheck=false}

    \vspace{-0.5em}
    \figcaption{Performance-memory plot on the UCF-101~\cite{soomro2012ucf101} dataset from the TCD benchmark~\cite{tcd}}{
\ours{} consistently outperforms existing methods across various memory sizes, achieving a superior performance-memory-efficiency trade-off.
    }

    \figcapmargin
    \label{fig:tradeoff}
\end{figure}

%% file: figure/fig_motivation.tex
\begin{figure}[t]
    \centering\centering
    \includegraphics[width=0.95\linewidth]{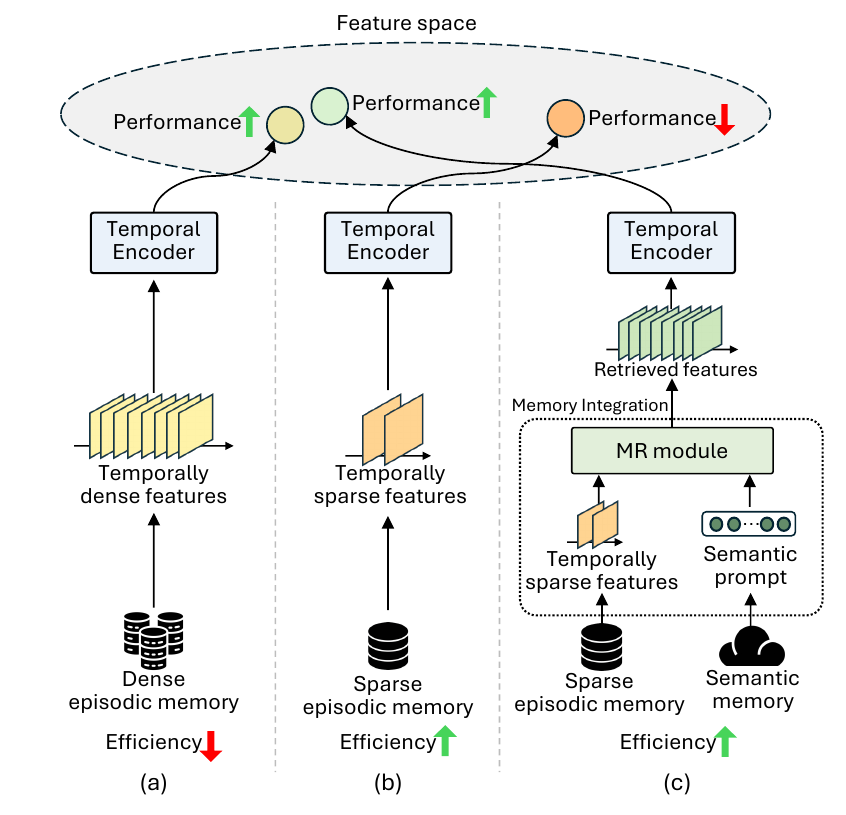}
\captionsetup{justification=justified, singlelinecheck=false}
    

    \vspace{-0.5em}
    \figcaption{MR module achieves a better performance-memory-efficiency trade-off}{
 (a) Storing temporally dense features in episodic memory achieves high performance but results in low memory-efficiency.
 %
 (b) Storing temporally sparse features improves memory-efficiency but sacrifices performance due to the lack of temporal context. 
(c) \ours{} tries to combine the best of both worlds. 
We only store temporally sparse features and lightweight prompts to maintain high memory-efficiency. 
The MR module integrates episodic and semantic memory to retrieve temporally dense features. 
The distance between the retrieved feature vector and the original temporally dense feature vector is significantly smaller than the distance between the temporally sparse feature vector and the temporally dense feature vector.
\ours{} uses these high-quality retrieved features for rehearsal training to mitigate catastrophic forgetting with high memory-efficiency.
}

    \figcapmargin
    \label{fig:motivation}
\end{figure}

%% file: 2_prior.tex
\section{Related Work}
\label{sec:related}
\vspace{-0.2em}
\paragraph{Action recognition.} 
There has been remarkable progress in video action recognition with deep learning over the past decade.
Early approaches primarily use CNNs, including 2D~\cite{donahue2016longterm,karpathy20142dcnn,lin2019tsm,Simonyan-NIPS-2014,ng2015short,zhou2018trn,Simonyan-NIPS-2014,feichtenhofer2019slowfast}, 3D~\cite{i3d,feichtenhofer2019slowfast,ji20123d,tran2015c3d,tran2018closer,wang2018non,feichtenhofer2020x3d} and (2+1)D separable CNNs~\cite{tran2018closer, Xie-ECCV-2018}. 
More recently, transformer architectures have shown significant performance improvements due to their global reasoning capabilities~\cite{arnab2021vivit, timesformer, motionformer, wu2022memvit, fan2021multiscale, yan2022multiview,girdhar2022omnivore,standalone} and the transferability of pre-trained knowledge~\cite{pan2022st,yang2023aim,lee2024cast}. 
Despite the great success, a common limitation of prior works is the tendency for models to lose knowledge learned from previous data when incrementally trained with new data. 
To address this limitation, we focus on the problem of video class-incremental learning in this work.


\vspace{-1em}
\paragraph{Class-incremental learning (CIL).} 
The main challenge in CIL is mitigating catastrophic forgetting.
%
%
Recent CIL approaches primarily focus on either rehearsal memory or parameter-efficient tuning to mitigate catastrophic forgetting of large pre-trained models.
Rehearsal-based CIL methods~\cite{rebuffi2017icarl,wu2019large,prabhu2020gdumb,wang2021triple,buzzega2020dark} store a few training samples per class for previous tasks in a rehearsal memory, and then use them as training data in the current task to retain previously learned knowledge. 
On the other hand, parameter-efficient tuning methods~\cite{wang2022learningl2p,wang2022s,wang2022dualprompt,smith2023coda,gao2023unified,tan2024semantically} employ a frozen backbone encoder pre-trained on a large-scale dataset to prevent catastrophic forgetting and learn a few parameters such as adapters or prompts to capture task-specific knowledge.
While both types of methods demonstrate impressive performance improvements, they are mostly focused on the image domain. 
In contrast, we address the relatively under-explored area of video class-incremental learning.


\vspace{-1em}
\paragraph{Video class-incremental learning (VCIL).} 

Compared to image CIL, VCIL is relatively under-explored due to the challenges of understanding temporal details essential for recognizing actions.
Recent VCIL approaches address these challenges mainly through regularization and prompt-based methods.
Regularization techniques~\cite{tcd,chengstsp,zhao2021video,villa2022vclimb,liang2024hypercorrelation} retain temporal information of previous tasks to reduce forgetting, while prompt-based methods~\cite{villa2023pivot,stprompt} leverage CLIP~\cite{clip} to combine visual and text features with learnable prompts to mitigate forgetting. 
A limitation of the prior methods is memory-inefficiency, as they store temporally dense data in episodic memory to preserve temporal details.
Storing temporally sparse data in episodic memory~\cite{alssum2023just,framemaker} could address memory concerns, but this approach often results in degraded performance due to insufficient temporal information.
In contrast, our approach achieves a better trade-off between memory-efficiency and performance by storing only temporally sparse features in episodic memory and using a memory retrieval (MR) module to retrieve temporally dense features during rehearsal stage.



%% file: 4_method.tex
\section{\ours{}}
\label{sec:method}
\input{figure/fig_overview}
\vspace{-0.5em}

%
%
Here, we first outline the core philosophy of \ours{} to achieve a better trade-off between memory-efficiency and performance in video class-incremental learning.
i) To reduce memory consumption, we store only \emph{temporally sparse} features in an episodic memory instead of \emph{temporally dense} features, and \emph{lightweight} semantic prompts.
ii) To effectively mitigate catastrophic forgetting and achieve high performance, the MR module retrieves \emph{temporally dense} features during rehearsal stage by applying cross-attention between \emph{temporally sparse} features and semantic prompts.
iii) To enable effective retrieval, we train the MR module during each incremental training stage to retrieve temporally dense features using temporally sparse features and semantic prompts as input.
%
%

%
In the following sections, we provide detailed description of \ours{}.
We begin by formulating the VCIL problem in \secref{formulation}, 
followed by details of incremental training stage of the proposed method in \secref{overview}. 
We provide detailed descriptions of the MR module in \secref{mr} and \secref{mr_learn}, and then describe rehearsal stage in \secref{rehearsal} and the inference process in \secref{inference}.

\subsection{Formulation}
\label{sec:formulation}

\vspace{-0.5em}
The goal of VCIL is to incrementally learn a model $F(\cdot; \theta, \phi) = h(f(\cdot; \theta); \phi)$ parameterized by $\theta$ and $\phi$ for sequentially introduced $K$ tasks $\mathbb{T}=\{T_1, T_2, \ldots, T_K\}$. 
Here, $f(\cdot; \theta)$ denotes a feature extractor, and $h(\cdot; \phi)$ is a classifier. 
For a task $T_k$, we have a dataset with $N_k$ training samples, $\SD_k = \{(\X_i^k, y_i^k)\}_{i=1}^{N_k}$, where $\X_i^k$ denotes the $i$-th video clip in the dataset, and $y_i^k$ is the corresponding video label belonging to the label set $\mathbb{Y}_k$.
We uniformly sample $L$ frames from an input video to obtain a clip
$\X_i^k= [\x_{i,1}^k, \x_{i,2}^k, \hdots \x_{i,L}^k] \in \mathbb{R}^{L \times H \times W \times C}$
, where $H$, $W$, and $C$ are the height, width, and the number of color channels, respectively.
In the VCIL setting, $\mathbb{Y}_k$ does not overlap with label sets of the other tasks. 

\vspace{-1.3em}
\paragraph{Memory.}
To mitigate catastrophic forgetting, \emph{episodic} memory is widely used in CIL~\cite{villa2023pivot,tcd,framemaker,wang2022learningl2p}. 
%
%
For the current task $T_k$, we train the model $F$ on the dataset $\SD_k$. 
Once incremental training stage $k$ is finished, we sample a subset of data from $\SD_k$ to store in episodic memory $\episodic_k$ for rehearsal stage, {following~\cite{villa2022vclimb,tcd}}.
We denote the episodic memory containing data from all previous tasks as $\episodic_{1:k} = \episodic_1 \cup \episodic_2 \cup \cdots \cup \episodic_{k}$. 
During rehearsal stage, we train the model $F$ on $\episodic_{1:k}$ to retain previously learned knowledge.
%
To complement episodic memory,
%
we introduce a learnable \emph{semantic prompt} $\P^k \in \mathbb{R}^{L \times d}$, which captures general knowledge within each task.
%
We store the semantic prompt in semantic memory $\semantic_k$.
%

\vspace{-0.5em}
\subsection{Incremental Training Stage}
\vspace{-0.5em}
\label{sec:overview}
For brevity, we omit the task index $k$ and the sample index $i$ unless explicitly stated.
As shown in \figref{overview} (a), given a video clip $\X$ of the current task dataset $\SD$, we extract temporally dense frame-level features 
$\S_{\text{dense}}=f_s(\X) \in \mathbb{R}^{L \times d}$, \ie \cls{} tokens, using a \emph{frozen} visual encoder $f_s(\cdot)$.
%
Then we pass the original temporally dense features $\S_{\text{dense}}$ into a \emph{learnable} temporal encoder $f_t(\cdot; \theta_t)$ to extract a clip-level feature vector $\z_{\text{temporal}} \in \mathbb{R}^{d}$ as follows:
 \begin{equation}
     \z_{\text{temporal}} = f_t(\S_{\text{dense}}; \theta_t).
     \label{eq:temp_feat}
 \end{equation}

\vspace{-2em}
\paragraph{Action classification loss.} 
The classifier $h(\cdot;\phi)$ takes the original clip-level feature vector $\z_\text{temporal}$ and predicts the action label $\hat{\y}$ as:
\vspace{-0.5em}
\begin{equation}
     \hat{\y} = h(\z_{\text{temporal}}; \phi).
     \label{eq:head}
\end{equation}
%
We employ the standard cross-entropy loss between the action prediction $\hat{\y}$ and the ground truth label $\y \in \mathbb{Y}$ from the current task training dataset $\SD$:
\begin{equation}
    {L}_{\text{CE}} = -\sum_{(\X, \y) \in \SD} \y \log(\hat{\y}).
    \label{eq:cur_ce}
\end{equation}

\paramargin
\vspace{-0.5em}
\subsubsection{Memory Retrieval Module}
\label{sec:mr}

The MR module is the key to achieving a high memory-efficiency-performance trade-off.
As shown in \figref{overview} (b) and (c), the MR module \emph{retrieves} temporally \emph{dense} features by integrating temporally \emph{sparse} features stored in episodic memory and the semantic prompt stored in semantic memory through cross-attention.
The features retrieved by the MR module are temporally enhanced 
compared to temporally sparse features, making them more effective in mitigating forgetting during rehearsal stage. 
%
%
Notice that, to prevent the MR module from catastrophic forgetting, we employ a dedicated MR module for each task.
We analyze the effectiveness of the MR module and semantic memory in \secref{analysis} and \secref{ablation}.

\vspace{-1em}
\paragraph{Integration of episodic and semantic memory.}
%
%
As shown in \figref{overview} (d), the MR module $\Phi(\cdot)$ \emph{retrieves} temporally dense features $\tilde{\S}_\text{dense} \in \mathbb{R}^{L \times d}$ through cross-attention between temporally sparse features $\S_{\text{sparse}}  \in \mathbb{R}^{l \times d}$ ($ l \ll L $) and the semantic prompt $\P \in \mathbb{R}^{L \times d}$ shared in the current task.
We uniformly subsample the temporally dense features $\S_{\text{dense}} $ to obtain the temporally sparse features $ \S_{\text{sparse}}$.
We can define the Multi-Head Cross-Attention ($\mhca$) from $\E_2\in\mathbb{R}^{N\times d}$ to $\E_1\in\mathbb{R}^{M\times d}$ as follows:
\vspace{-0.5em}
\begin{equation}
\mhca(\E_1, \E_2) = \softmax(\frac{(\E_1 \mathbf{W_Q})(\E_2 \mathbf{W_K})^\top}{\sqrt{d}})(\E_2 \mathbf{W_V}),
\end{equation}
where $\mathbf{W_Q},\mathbf{W_K}$, and $\mathbf{W_V}$ are the learnable query, key, and value projection matrices, respectively.
In the MR module the semantic prompt $\P$ serves as the query while the temporally sparse features $\S_{\text{sparse}}$ serves as the key and value.
We apply the cross-attention operation along with a residual connection:
\vspace{-0.5em}
\begin{equation}
\S^\prime = \mhca(\lnorm(\P ), \lnorm(\S_{\text{sparse}})) + \P ,
\end{equation}
where $\lnorm(\cdot)$ denotes the layer normalization operation~\cite{layernorm}. 
Finally, we obtain the final output of the MR module by feeding $\S^\prime$ into a feed-forward network ($\ffn$)~\cite{vit} with a residual connection.
\vspace{-0.5em}
\begin{equation}
\tilde{\S}_{\text{dense}}=\Phi (\P , \S_{\text{sparse}}) = \ffn(\lnorm(\S^\prime))+\S^\prime.
\end{equation}
%
%
Then we can feed the MR module retrieved features $\tilde{\S}_{\text{dense}}$ into the temporal encoder $f_t(\cdot;\theta_t)$, to obtain a clip-level feature vector as follows:
\vspace{-0.5em}
\begin{equation}
\tilde{\z}_\text{temporal}=f_t(\tilde{\S}_{\text{dense}};\theta_t).
\label{eq:z_tilde}
\end{equation}

\subsubsection{MR Module Training}
\label{sec:mr_learn}
To enable effective retrieval, we train the MR module using the following loss terms.

\vspace{-1.2em}
\paragraph{Static matching loss.}
We aim for the MR module to retrieve temporally dense features $\tilde{\S}_{\text{dense}}$ that closely resemble the original temporally dense features $\S_{\text{dense}}$.
To achieve this, we apply a frame-level $L_2$ loss between them:
\vspace{-0.5em}
\begin{equation}
     L_{\text{SM}} = \left\|\tilde{\S}_{\text{dense}} - \S_{\text{dense}} \right\|_2^2.
     \label{eq:sm}
\end{equation}

\vspace{-1em}
\paramargin

\paragraph{Temporal matching loss.}

We also aim for the clip-level feature vector \(\tilde{\z}_{\text{temporal}}\) obtained using \eqnref{z_tilde}, to closely resemble the original clip-level feature vector $\mathbf{z}_\text{temporal}$ obtained by \eqnref{temp_feat}.
To achieve this, we apply an $L_2$ loss between them:
\vspace{-0.5em}
\begin{equation}
    L_{\text{TM}} = \left\|\tilde{\z}_{\text{temporal}} - \z_{\text{temporal}} \right\|_2^2.
    \label{eq:tm}
\end{equation}
%

\vspace{-1em}
\paragraph{Total loss.}
We define the total loss function for the current task training as follows:
\vspace{-0.5em}
\begin{equation}
L_{\text{total}} =  L_{\text{CE}} + \alpha  L_{\text{SM}} + \beta L_{\text{TM}},
    \label{eq:total}
\end{equation}
where $\alpha$ and $\beta$ are weighting hyperparameters.
Backpropagating the gradient of the total loss function enables learning not only for the classifier and temporal encoder but also for the retrieval capability of the MR module.
Additionally, we posit that the semantic prompt captures some \emph{general} knowledge across different samples and classes within the current incremental training stage. 
We analyze the general knowledge captured by semantic prompts in \secref{analysis}.
We empirically find that using both the MR module and the semantic prompt is effective in VCIL (\secref{ablation}).

\vspace{-1em}
\paragraph{Memory write.}
Once an incremental training stage $k$ is finished, we store temporally sparse features $\S_{\text{sparse}}$ in the episodic memory following prior works~\cite{villa2022vclimb,tcd}, and the learned semantic prompt $\prompt^k$ in the semantic memory.
\vspace{-0.5em}
\begin{equation}
    \SM^E_k = \{\S_{\text{sparse}}^i \mid \forall i \in \mathbb{I}\}, \quad
    \SM^S_k = \prompt^k.
    \label{eq:mem_write}
\end{equation}
Here, $\mathbb{I}$ denotes the set of selected sample indices.
The combined memory for the current task is $\SM_k = \SM_k^E \cup \SM_k^S$.

\subsection{Rehearsal Training}
\label{sec:rehearsal}

As shown in \figref{overview} (c), during rehearsal stage, we train the temporal encoder $f_t(\cdot; \theta_t)$ and classifier $h(\cdot;\phi)$ using the temporally dense features retrieved by the \emph{frozen} MR module rather than temporally sparse features. 
%
Specifically, we select the corresponding MR module and the semantic prompt based on the task ID associated with the features stored in the episodic memory. 
We employ the standard cross-entropy loss using the clip-level feature vector $\tilde{\z}^k_{\text{temporal}}$ as follows:

\begin{equation}
     L_{\text{R}} = -\smashoperator{\sum_{(\X^k, \y^k) \in \SM_{1:k}}} \y^k \log(h(\tilde{\z}^k_{\text{temporal}}; \phi)).
    \label{eq:loss_mem}
\end{equation}

%
By leveraging the MR module to retrieve temporally dense features for rehearsal training while storing only temporally sparse features, we effectively address the challenge of limited temporal context inherent in temporally sparse features (\figref{semantic}). 
The proposed memory retrieval mechanism enables better preservation of temporal information from previous task videos, which is essential for video understanding.
%
As a result, \ours{} successfully mitigates forgetting while achieving high memory-efficiency.

\subsection{Inference}
\label{sec:inference}
\paramargin

During inference, we use the visual encoder, temporal encoder, and classifier, and drop the other components. 
Given an input video clip, we obtain an action prediction using \eqnref{temp_feat}, and \eqnref{head}.
Remarkably, \ours{} is more \emph{computation-efficient} during inference compared to prior works~\cite{villa2023pivot,stprompt}, which require two forward passes due to a query function. 

%% file: figure/fig_overview.tex
\begin{figure*}[t]
    \centering
    \includegraphics[width=\linewidth]{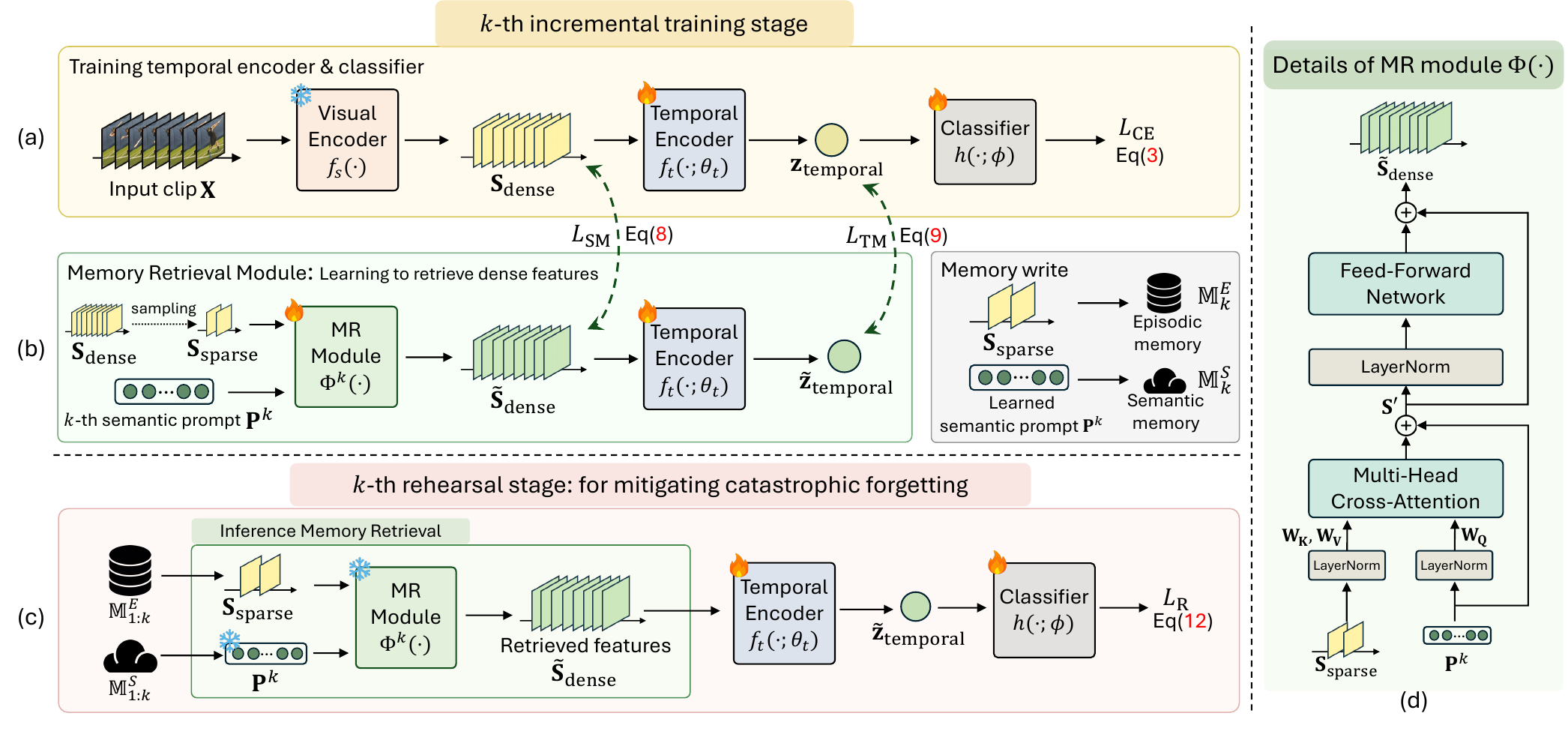}

    \vspace{-0.5em}
    \figcaption{Overview of \ours{}}{
    (a) \ours{} employs a frozen visual encoder to extract frame-level features $\S_{\text{dense}}$ from the input clip.
    We pass the frame-level features into a learnable temporal encoder to extract a clip-level feature vector $\z_{\text{temporal}}$.
    (b) We introduce a memory retrieval (MR) module designed to retrieve temporally dense features from temporally sparse features. 
    To enable effective retrieval, we employ the static and temporal matching losses for MR module training. 
    (c) During the $k$-th rehearsal stage, the MR module integrates episodic memory and semantic memory through cross-attention, retrieving temporally dense features from temporally sparse features.
    \ours{} leverages retrieved features for rehearsal training, effectively mitigating catastrophic forgetting with high memory-efficiency
    (d) The MR module consists of a cross-attention mechanism between the learnable semantic prompt and temporally sparse features denoted as $\S_\text{sparse}$. 
    %
    Through MR module training, the semantic prompt learns general knowledge, and the MR module learns to retrieve the temporally dense features using only temporally sparse features and the semantic prompt.
        %
}
\label{fig:overview}
\end{figure*}

    
    
    

%% file: 5_results.tex
\section{Experimental Results}
\label{sec:results}
\input{table/vclimb}

\vspace{-0.5em}
We conduct experiments to answer the following research questions: 
(1) Does the MR module effectively retrieve temporally dense features? (\secref{analysis})
(2) Is \ours{} robust to a decreasing number of frame features (\secref{analysis})
(3) Do the semantic prompts learn general knowledge? (\secref{analysis})
(4) What are the key components for effective VCIL? (\secref{ablation})
(5) Does \ours{} outperform existing methods? (\secref{main_table}).
To this end, we first describe the datasets (\secref{dataset}) and the evaluation metrics (\secref{metric}) used.
We put the implementation details in the supplementary material.

\subsection{Datasets}
\label{sec:dataset}

We conduct the experiments on four static-biased~\cite{li2018resound,choi2019can,sevilla2021only} datasets UCF-101~\cite{soomro2012ucf101}, HMDB51~\cite{hmdb}, ActivityNet~\cite{caba2015activitynet}, and Kinetics-400~\cite{kinetics}, and one temporal-biased~\cite{timesformer,kowal2022deeper} dataset, Something-Something-V2 (SSV2)~\cite{ssv2}.
%
%
We utilize dataset splits from two benchmarks. \romannumeral 1) vCLIMB benchmark~\cite{villa2022vclimb} includes UCF-101, ActivityNet, and Kinetics-400. 
\romannumeral 2) TCD benchmark~\cite{tcd} includes UCF-101, HMDB51, and SSV2.
Please refer to the supplementary material for more details.


%

\subsection{Evaluation Metric}
\label{sec:metric}

\paramargin
\paragraph{Performance.}

For the vCLIMB~\cite{villa2022vclimb} benchmark, report Average Accuracy (AA). 
For the TCD~\cite{tcd} benchmark, we report Average Incremental Accuracy (AIA). 
For more details, please refer to the supplementary material.

\vspace{-1em}
\paragraph{Memory usage.}
We measure the memory usage for episodic memory by $N_s \times l \times d \times 4$, where $N_s$ is the number of samples per class, $l$ is the temporal length of the frame features, and $d$ is the frame-level feature vector dimension.
For semantic memory, we calculate the memory usage as $L \times d \times 4$ per task.
We assume each feature vector requires 4 bytes.
%
We provide detailed memory usage calculations for both existing methods and \ours{} (including the MR module) in the supplementary material.

\subsection{Comparison with State-of-the-Art}
\label{sec:main_table}
\paragraph{vCLIMB benchmark~\cite{villa2022vclimb}: UCF-101, ActivityNet, and Kinetics-400 results.}
In \tabref{vclimb}, we evaluate the proposed and existing methods.
%
%
Compared to ST-Prompt~\cite{stprompt} and PIVOT~\cite{villa2023pivot}, which utilize the CLIP~\cite{clip} backbone with learnable prompts, \ours{} achieves significantly higher performance with substantially lower memory consumption.  
%
Similar to ours, SMILE~\cite{alssum2023just} stores temporally sparse data in episodic memory.
However, SMILE consumes significantly higher memory than ours as it stores a larger number of samples.
%
%
In summary, \ours{} achieves state-of-the-art performance and memory-efficiency on all datasets of the vCLIMB.

\input{table/tcd}

\vspace{-1em}
\paragraph{TCD benchmark~\cite{tcd}: UCF-101, HMDB51, and SSV2 results.}
In \tabref{tcd}, we compare \ours{} with existing methods.
%
%
Compared to existing methods, \ours{} shows the best performance in 5 out of 7 settings with the highest memory-efficiency in all settings.
%
Specifically, compared to ST-Prompt~\cite{stprompt}, \ours{} achieves superior performance in all tasks with the same backbone while using less than 6\% of the memory.  
Although STSP~\cite{chengstsp} shows strong performance on the SSV2 dataset, it consumes approximately $\times 67.9$ more memory than \ours{}. 
%
 %
%
These results demonstrate that \ours{} achieves a better trade-off between performance and memory-efficiency compared to existing methods.

\input{figure/frame_robust}

\subsection{Analysis}
\label{sec:analysis}



\paragraph{Is \ours{} robust to a decreasing number of frame features?}

In \figref{robust}, we examine the robustness of \ours{} to a decreasing number of the frame features $l$ stored in the episodic memory.
Here, we fix the number of stored samples per class as $N_s=4$.
%
The baseline, which relies solely on episodic memory without semantic memory and the MR module, shows a significant performance drop as $l$ decreases (dashed lines). 
In contrast, our method shows a much more moderate decline in performance (solid lines). 
The results indicate that \ours{} effectively mitigates forgetting even when storing temporally sparse features, thanks to the retrieval capability of the MR module.
This retrieval enables us to store only temporally sparse features in episodic memory, leading to high memory-efficiency.

%


\vspace{-1.0em}
\paragraph{Does the MR module effectively retrieve temporally dense features?}
To evaluate the retrieval capability of the MR module, we compare the $L^2$ distances from the original temporally dense feature vectors ($\S_{\text{dense}}$) to i) temporally sparse feature vectors ($\S_{\text{sparse}}$) and ii) retrieved feature vectors ($\tilde{\S}_{\text{dense}}$).
We match the temporal dimension of $\S_{\text{sparse}}$ and $\S_{\text{dense}}$ by upsampling $\S_{\text{sparse}}$ with the nearest neighbor method. 
As shown in \figref{semantic}, we observe that the distance between $\S_{\text{dense}}$ and $\tilde{\S}_{\text{dense}}$ is reduced compared to the distance between $\S_{\text{dense}}$ and $\S_{\text{sparse}}$ ($8.2$ \vs $19.4$).
We can qualitatively observe that the retrieved feature vectors, ($\tilde{\S}_{\text{dense}}$, \recon{circles}), 
are closer to the original temporally dense feature vectors, ($\S_{\text{dense}}$, \densec{crosses}), compared to the sparse feature vectors, ($\S_{\text{sparse}}$, \sparsec{squares}).
This result indicates that the MR module effectively retrieves temporally dense features from temporally sparse features.

\input{figure/fig_semantic}

\vspace{-1.3em}
\paragraph{Do the semantic prompts learn general knowledge?}
To investigate whether the semantic prompts capture general knowledge, we compare retrieval performance using (i) learned semantic prompt and (ii) randomly initialized semantic prompt.
Specifically, we measure the $L_2$ distance and visualize the vectors using T-SNE~\cite{van2008visualizingtsne} for feature vectors retrieved with the learned semantic prompt, ($\tilde{\S}_{\text{dense}}$, \recon{circles}) versus feature vectors retrieved with randomly initialized semantic prompt, ($\tilde{\S}^{\text{random}}_{\text{dense}}$, \random{circles}). 
As shown in \figref{semantic}, retrieval is significantly more effective with learned semantic prompt compared to using a randomly initialized semantic prompt ($8.2$ \vs $33.2$).
\ours{} can effectively retrieve temporally dense feature vectors belonging to multiple classes using a \emph{single} prompt.
Both the quantitative evaluation and visualization demonstrate that the semantic prompt learns \emph{general knowledge}.

\input{table/ablation}


\subsection{Ablation Study}
\label{sec:ablation}
\vspace{-0.3em}

We conduct extensive ablation studies to examine the design choices of the proposed method on the SSV2 dataset. 
%

\paramargin

\vspace{-1em}
\paragraph{Effect of type of MR module and semantic memory.}
In \tabref{abl} (a), we conduct experiments to examine the effect of different configurations of the MR module and semantic memory. 
%
%
We can design both the MR module and semantic memory in a global, task-specific, or class-specific manner.
In \tabref{abl} (a), the \emph{Global} configuration refers to a single component shared across all tasks, while the \emph{Task} and \emph{Class} configurations denote components shared within each task and each class, respectively.
%
%
The first row represents a baseline using only episodic memory without the MR module and semantic memory, achieving 42.1\% accuracy.
When we introduce the global MR module and global semantic memory, we observe a 5.3-point increase in accuracy over the baseline, demonstrating the effectiveness of the MR module and semantic memory.
%
Using the global MR module with task-specific semantic memory results in a 0.7-point increase in accuracy compared to the global-global setup.
When both the MR module and semantic memory are task-specific, we achieve the highest accuracy of 48.9\%,  as the task-specific MR module prevents itself from forgetting.
Finally, when using a task-specific MR module with a class-specific semantic memory results in a 1.0-point decrease in accuracy. 
The result suggests that semantic memory is more effective for mitigating forgetting when it shares general knowledge at the task-level rather than the class-level.

\paragraph{Different memory retrieval methods.}
%
In \tabref{abl} (b), we analyze different architectures for the MR module.
%
In the \emph{Frame interpolation} baseline, we store temporally sparse RGB frames. 
During rehearsal stage, we employ an off-the-shelf frame interpolation method~\cite{reda2022film} to generate rgb frames, resulting in temporally dense rgb frames.
\emph{Interpolation} and \emph{MLP} refer to obtaining temporally dense features from \(\S_{\text{sparse}}\) without semantic memory, using linear interpolation and an MLP, respectively.
%
%
%
\emph{Add} and \emph{Multiply} denote simple element-wise addition and multiplication between \(\S_{\text{sparse}}\) and the semantic prompt \(\P\), respectively.
To match the temporal resolution $L$, we upsample $\S_{\text{sparse}}$ with the nearest neighbor method along the temporal axis.
%
%
Among all MR module architectures, cross-attention between semantic prompts and \(\S_{\text{sparse}}\) achieves the highest performance of 48.9\%, validating the design choice.
%

\vspace{-1em}
\paragraph{Effect of matching loss functions.}
In \tabref{abl} (c), we conduct an ablation study on the matching loss functions.
When used separately, both static and temporal matching losses give a slight improvement.
Employing both matching loss functions results in a significant performance improvement of 2.6 points.
Please refer to the supplementary material for more experiments on the balance between them.


\vspace{-1em}
\paragraph{Effect of semantic prompt length.}
In this experiment, we fix the clip length at $L = 8$ and examine the effect of varying prompt lengths, as shown in \tabref{abl} (d). 
As the prompt length increases from 8 to 24, we observe a corresponding improvement in accuracy, along with an increase in memory usage.
%
For prompt lengths of 16 and 24, it is important to note that we cannot apply the static matching loss in \cref{eq:sm} since the length of semantic prompt must match the clip length.
%
A prompt length of 8 with static matching loss achieves the highest accuracy at 48.9\% while being the most memory-efficient.
Therefore, we set the semantic prompt length equal to the clip length to achieve high memory-efficiency and performance.


\vspace{-1.3em}
\paragraph{Effect of frame sampling strategies for episodic memory storage.}
In \tabref{abl} (e), we investigate different frame sampling strategies for storing temporally sparse features in episodic memory.
The first row (random) presents a simple random selection along the temporal axis, achieving a performance of 48.2\%. 
In the second row, the temporal attention approach selects the top-$l$ frames with the highest attention scores from the temporal encoder $f_t(\cdot:,\theta)$,  resulting in a performance of 48.6\%. 
Lastly, uniformly sampling $l$ frames across the entire video, achieves the best performance at 48.9\%.
These results suggest that uniform sampling, which covers the entire video length, is most beneficial for effective memory retrieval.

\vspace{-1.3em}
\paragraph{Effect of input features to MR module during training stage.}
In \tabref{abl} (f), we examine the effect of different input features on the MR module during training stage.
We observe that using temporally sparse features \( \S_\text{sparse} \) as input yields a 1.3-point improvement over using temporally dense features \( \S_\text{dense} \) when training the MR module. 
During rehearsal stage, we aim to retrieve temporally dense features from temporally sparse features stored in episodic memory. 
Training the MR module with $\S_\text{sparse}$ as an input during training stage further enhances its retrieval capability for an effective rehearsal stage.


%% file: table/vclimb.tex
\begin{table*}[t]
    \centering
    \scriptsize
    \renewcommand{\arraystretch}{1.1}
    \setlength{\tabcolsep}{4pt}
    \caption{\textbf{Comparison with the state-of-the-arts on the vCLIMB Benchmark~\cite{villa2022vclimb}.} We report the Top-1 average accuracy (\%) and
the total memory usage (MiB). We indicate the backbone model in parentheses. The best are in \best{bold}, and the second best are \second{underscored}. A dash (-) denotes a value not reported in the original paper.
{\ours{} achieves the best performance with minimal memory consumption across all datasets in the benchmark.}
}
    \vspace{-1em}    
    
    \begin{tabular}{l r r r r r r r r r r r r r  }
        \toprule

        \multirow{4}{*}{Method} & 
        \multicolumn{4}{c}{ActivityNet}& 
        \multicolumn{4}{c}{Kinetics-400} & 
        \multicolumn{4}{c}{UCF-101} \\
            \cmidrule(lr){2-5} \cmidrule(lr){6-9} \cmidrule(lr){10-13} 

        & \multicolumn{2}{c}{10 Tasks}& 
        \multicolumn{2}{c}{20 Tasks}& 
        \multicolumn{2}{c}{10 Tasks}& 
        \multicolumn{2}{c}{20 Tasks}&
        \multicolumn{2}{c}{10 Tasks}&
        \multicolumn{2}{c}{20 Tasks} \\
        
    \cmidrule(lr){2-3}
    \cmidrule(lr){4-5}
    \cmidrule(lr){6-7}
    \cmidrule(lr){8-9}
    \cmidrule(lr){10-11}
    \cmidrule(lr){12-13}
         
                             & Acc. & MEM. & Acc. & MEM. & Acc. & MEM. & Acc. & MEM. & Acc. & MEM. & Acc. & MEM. \\
        \midrule
        vCLIMB+iCaRL~\cite{villa2022vclimb} (TSN)  & 48.5 & 2.2T & 43.3 & 2.2T  & 32.0 & 287.1G & 26.7 & 287.1G & 81.0 & 529.7G & 76.6 & 529.7G \\
        vCLIMB+BiC~\cite{villa2022vclimb} (TSN)    & 52.0 & 2.2T & 46.5 & 2.2T & 28.0 & 287.1G & 23.1 & 287.1G & 78.2 & 529.7G & 70.7 & 529.7G \\
        vCLIMB+iCaRL+TC~\cite{villa2022vclimb} (TSN)  & 44.0 & 9.1G & - & - & 36.5 & 18.3G & - & - & 75.8 & 4.6G & - & - \\
        SMILE+iCaRL~\cite{alssum2023just} (TSN)   & 50.3 & 2.2G & 43.5 & 2.2G & 46.6 & 35.3G & 45.8 & 35.3G & 95.7 & 1.3G & 95.9 & 1.3G \\
        SMILE+BiC~\cite{alssum2023just} (TSN)     & 54.8 & \second{2.2G} & 51.1 & \second{2.2G} & 52.2 & 35.3G & 48.2 & 35.3G & 92.5 & 1.3G & 90.9 & 1.3G \\
        ST-Prompt~\cite{stprompt} (CLIP)    & - & - & - & - & 54.6 & *\second{318.8M} & 54.5 & *\second{318.8M} & 85.1 & *\second{80.5M} & 85.3 & *\second{80.5M} \\
        PIVOT~\cite{villa2023pivot} (CLIP)        & \second{73.8} & *1.1T & \second{73.8} & *1.1T & \second{55.1} & *143.5G & \second{55.0} & *143.5G & \second{93.4} & *26.4G & \second{93.1} & *26.4G \\
        \midrule 
        \ours{} (ViT)     & 73.9 & 52.8M& 72.8 & 52.8M & 53.8 & 112.8M & 53.5 & 112.8M & 91.5 & 14.6M & 91.2 & 14.6M \\
        \ours{} (CLIP)     & \best{77.3} & \best{37.7M} & \best{75.1} & \best{37.9M} & \best{58.8} & \best{75.2M} & \best{58.5} & \best{75.5M} & \best{95.8} & \best{9.7M} & \best{95.9} & \best{9.9M} \\
        \bottomrule
     *vCLIMB and PIVOT store all frames of each stored video.
    \end{tabular}
    \label{tab:vclimb}
    \vspace{-1.5em}
\end{table*}


%% file: table/tcd.tex
\begin{table*}[t]
    \centering
    \scriptsize
    \renewcommand{\arraystretch}{1.2}
    \setlength{\tabcolsep}{4pt}

    \caption{\textbf{Comparison with the state-of-the-arts on the TCD Benchmark~\cite{tcd}.} We report the Top-1 average incremental accuracy (\%) and the total memory usage (MiB). We indicate the backbone model in parentheses. An asterisk (*) denotes estimated memory usage. The best are in \best{bold} and the second best are \second{underscored}. 
    }
    \vspace{-1em}
    
    \begin{tabular}{l  r r r r  r r r r  r r  r r  r r }
        \toprule
         &
         \multicolumn{6}{c}{UCF-101} &
         \multicolumn{4}{c}{HMDB51} &
         \multicolumn{4}{c}{SSV2} \\
        \cmidrule(lr){2-7} \cmidrule(lr){8-11} \cmidrule(lr){12-15} 
         
        Method & 
        \multicolumn{2}{c}{10 $\times$ 5 Tasks} & 
        \multicolumn{2}{c}{5 $\times$ 10 Tasks} & 
        \multicolumn{2}{c}{2 $\times$ 25 Tasks} & 
        \multicolumn{2}{c}{5 $\times$ 5 Tasks}  &
        \multicolumn{2}{c}{1 $\times$ 25 Tasks} & 
        \multicolumn{2}{c}{10 $\times$ 9 Tasks} &
        \multicolumn{2}{c}{5 $\times$ 18 Tasks} \\
        \cmidrule(lr){2-3} \cmidrule(lr){4-5} \cmidrule(lr){6-7} \cmidrule(lr){8-9} \cmidrule(lr){10-11} \cmidrule(lr){12-13} \cmidrule(lr){14-15}
         
        &Acc. & MEM. &Acc. & MEM. &Acc. & MEM. &Acc. & MEM. &Acc. & MEM. &Acc. & MEM. &Acc. & MEM. \\
        \midrule
        iCaRL~\cite{rebuffi2017icarl} (ViT)     & 70.6 & 1159.9M & 69.5 & 1159.9M & 67.3 & 1159.9M & 43.9 & 585.7M & 37.2 & 585.7M & 20.4 & 999.1M & 16.6 & 999.1M  \\
        UCIR~\cite{hou2019learningUCIR} (ViT)   & 77.6 & 1159.9M & 74.6 & 1159.9M & 71.8 & 1159.9M & 48.2 & 585.7M & 39.4 & 585.7M & 24.3 & 999.1M & 19.3 & 999.1M  \\
        L2P~\cite{wang2022learningl2p} (ViT)    & 81.2 &1159.9M  & 80.1 &1159.9M& 78.6 &1159.9M& 50.0 &585.7M& 45.9 &585.7M& 26.0 &999.1M& 21.3&999.1M\\
        TCD~\cite{tcd} (ViT)                    & 78.1 &1159.9M & 76.9 &1159.9M  &75.7&1159.9M &  50.3 &585.7M& 44.0 &585.7M& 29.3 & 999.1M& 24.7& 999.1M\\         
        TCD~\cite{tcd} (TSM)                    & 77.2 & 580.0M  & 75.4 & 580.0M  & 74.0 & 580.0M  & 50.4 & 292.8M & 46.7 & 292.8M & 35.8 & 4.0G & 29.6 & 4.0G    \\ 

        FrameMaker~\cite{framemaker} (TSM)      & 78.6 & 580.0M  & 78.1 & 580.0M  & 77.5 & 580.0M  & 51.1 & 292.8M &47.4 & 292.8M &37.3 & *4.0G& 31.0 & *4.0G \\ 
        
        FrameMaker~\cite{framemaker} (ViT)      &79.4  & 1159.9M & 79.6 &1159.9M& 79.3&1159.9M& 51.4&585.7M& 46.4&585.7M& 31.4 &999.1M&26.6&999.1M\\   
        
        ST-prompt~\cite{stprompt} (ViT)         & 83.7 & *86.7M & 81.5& *86.7M & 82.7 &*86.7M& 55.1 & *43.8M & 55.4 &*43.8M &  36.8 &*149.4M& 31.6&*149.4M \\
        
        ST-prompt~\cite{stprompt} (CLIP)        & \second{84.8} &*\second{80.5M} & \second{85.5}& *\second{80.5M} & \second{85.7}&*\second{80.5M} &  \second{60.1}& *\second{40.6M}& \second{60.5}&*\second{40.6M} & 40.0& *\second{138.7M}& 35.4&*\second{138.7M} \\
        
        STSP~\cite{chengstsp} (TSM)             & 81.2 &* 732.0M& 82.8 &* 732.0M& 79.3 &* 732.0M& 57.0 &*570.0M& 49.2 &*570.0M& \best{69.7} &*570.0M& \best{70.9} &*570.0M \\
    HCE~\cite{liang2024hypercorrelation}(TSM)   & 80.0 & 580.0M  & 78.8 & 580.0M  & 77.6 & 580.0M  & 52.0 & 292.8M & 48.9 & 292.8M & 38.7 & 4.0G & 32.8 & 4.0G  \\
        
        \midrule 
        
        \ours{} (ViT)     & 92.8 & 3.1M  & 90.8 & 3.2M  & 90.1 & 3.6M  & 62.3 & 1.6M  & 60.4 & 2.1M  & 44.9 & 8.4M  & 42.7 & 8.6M      \\
        
        \ours{} (CLIP)     &\textbf{95.1} & \textbf{3.1M}  & \textbf{93.9} & \textbf{3.2M}  & \textbf{93.3} & \textbf{3.6M}  & \textbf{65.7}&\textbf{1.6M}&\textbf{62.4} & \textbf{2.1M}& \second{48.9} & \textbf{8.4M}  & \second{47.5} & \textbf{8.6M}\\
        \bottomrule
    \end{tabular}
    \label{tab:tcd}
    \vspace{-1.5em}
\end{table*}

%% file: figure/frame_robust.tex
\begin{figure}[t]
\centering
\includegraphics[width=0.95\linewidth]{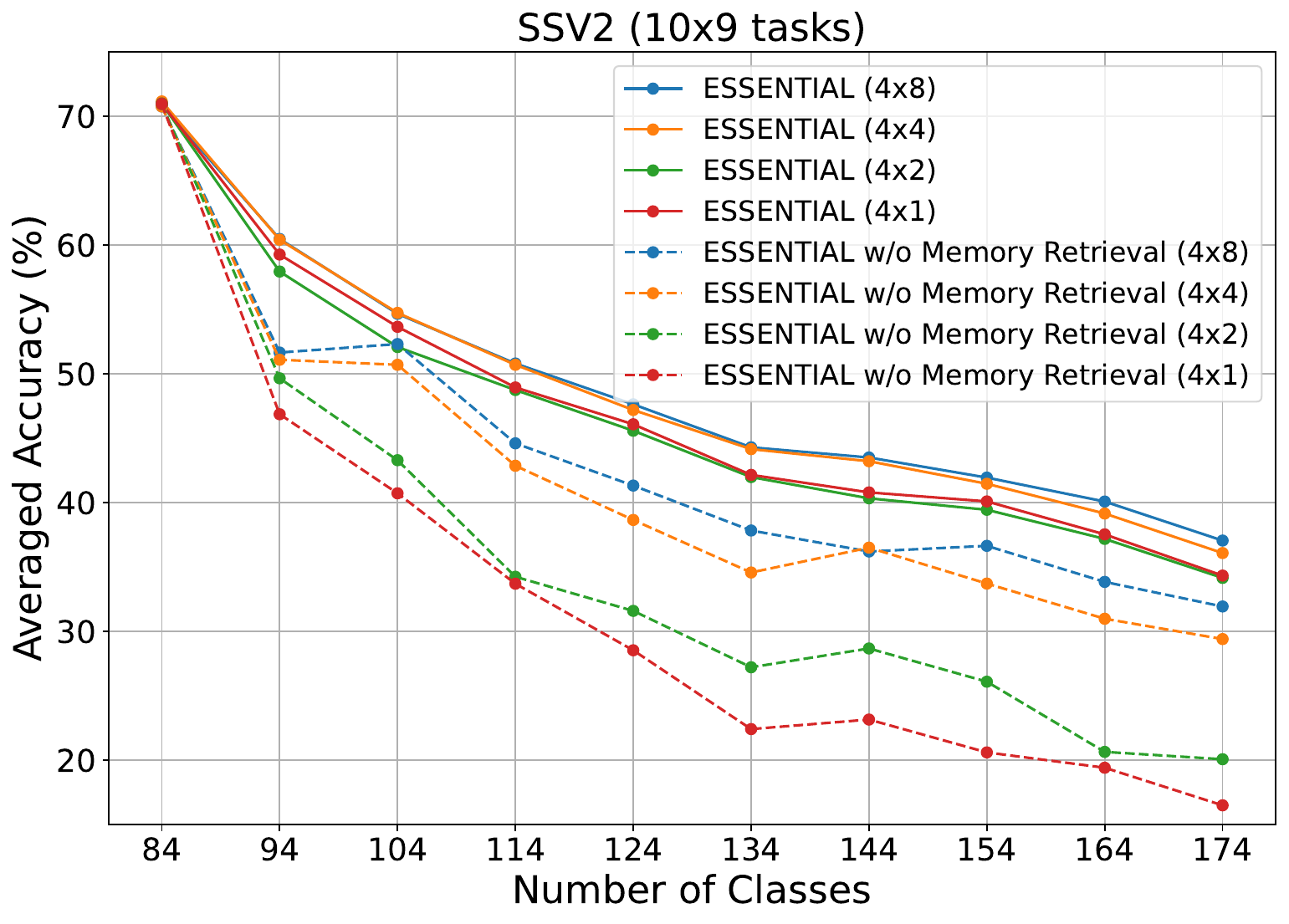}
    \vspace{-1em}
      \caption{\textbf{\ours{} is robust to a decreasing number of frame features.} 
        We conduct experiments on the SSV2 ($10 \times 9$ tasks) dataset. 
        Dashed lines are baseline relying solely on episodic memory and solid lines are \ours{} with different numbers of frame features stored in episodic memory.
        }
    \vspace{-2em}
\label{fig:robust}
\end{figure}


%% file: figure/fig_semantic.tex
\begin{figure}[t]
    \centering
    \includegraphics[width=\linewidth]{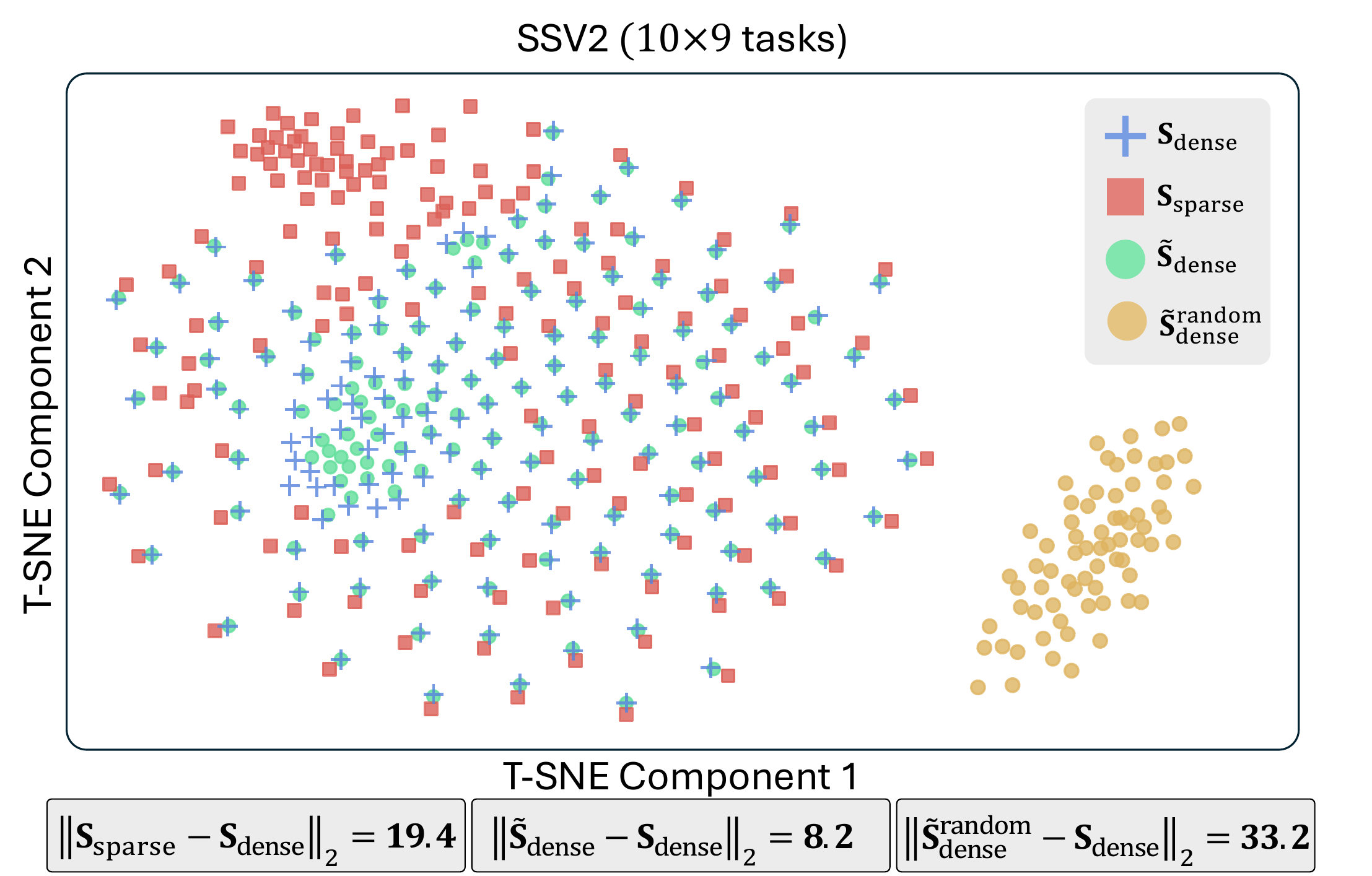}

    \caption{\textbf{The MR module effectively retrieves temporally dense features by leveraging general knowledge.}
    We measure the $L^2$ distances from the original temporally dense feature vectors to (a) temporally sparse feature vectors and (b) retrieved feature vectors with 
    i) learned semantic prompt 
    and
    ii) randomly initialized semantic prompt.
    We visualize the vectors of all classes from the seventh task of SSV2 ($10\times9$ tasks) using T-SNE~\cite{van2008visualizingtsne}.
    }
    \label{fig:semantic}
    \vspace{-1.5em}
\end{figure}


%% file: table/ablation.tex
\begin{table*}[t]
\centering
\caption{\textbf{Ablation study.} 
To validate the effect of each component, we show the results on the SSV2 
(10 $\times$ 9 tasks) dataset. We report the Top-1 average incremental accuracy (\%). 
We use CLIP~\cite{clip} as a backbone encoder and set $N_s=4$, and $l=4$ for episodic memory. 
}
\vspace{-0.5em}
\mpage{0.32}{\fontsize{8}{10}\selectfont (a) Effect of MR and semantic memory.}\hfill
\mpage{0.32}{\fontsize{8}{10}\selectfont (b) Different memory retrieval methods.}\hfill
\mpage{0.32}{\fontsize{8}{10}\selectfont (c) Effect of matching losses.}\hfill
\\
\mpage{0.32}{
\resizebox{0.80\linewidth}{!}{
\begin{tabular}{ccc}
\toprule
\multicolumn{2}{c}{Strategy} & \multirow{2}{*}{Acc.}  \\
\cmidrule{0-1}
MR module & Semantic memory  & \\
\midrule
\ding{55} &\ding{55} &42.1\\
Global & Global &47.4\\
Global & Task &48.1\\
Task & Task & \best{48.9}\\
Task & Class & 47.9\\  



\bottomrule
\end{tabular}
}
}
\mpage{0.32}{
\resizebox{\linewidth}{!}{
\begin{tabular}{lccc}
\toprule
Strategy & Semantic memory & Acc. \\
\midrule
Frame interpolation &\ding{55} & 45.5 \\
Feature interpolation&\ding{55} & 43.2 \\
MLP& \ding{55} & 45.8\\
Add&\ding{51} &46.7  \\
Multiply&\ding{51} & 46.5  \\
Self-attention \& FFN &\ding{51}& 48.1 \\
Cross-attention \& FFN &\ding{51}& \best{48.9} \\
\bottomrule
\end{tabular}
}
}
\mpage{0.32}{
\resizebox{0.95\linewidth}{!}{
\begin{tabular}{cc c}
\toprule
\multicolumn{2}{c}{Loss} & \multirow{2}{*}{Acc.} \\
\cmidrule{0-1}
Static matching & Temporal matching &  \\
\midrule
\ding{55} & \ding{55} & 46.3 \\
\ding{51}  &\ding{55}  & 46.9  \\
\ding{55}  &\ding{51}  & 47.1  \\
\ding{51}  &\ding{51}  & \best{48.9} \\ 
\bottomrule


\vspace{-0.5em}
\end{tabular}
}
}

 


\hfill

\vspace{2mm}
\mpage{0.4}{\fontsize{8}{10}\selectfont (d) Effect of semantic prompt length.}\hfill
\mpage{0.28}{\fontsize{8}{10}\selectfont (e) Effect of frame sampling strategies for episodic memory storage.}\hfill
\mpage{0.28}{\fontsize{8}{10}\selectfont (f) Effect of input features to MR module during training stage.}\hfill 
\\
\mpage{0.4}{
\resizebox{0.75\linewidth}{!}{
\begin{tabular}{lccc}
\toprule
Length & MEM.(MiB) &Static matching& Acc. \\
\midrule
8 & 0.23 & \ding{55}&47.1\\
16 & 0.47 & \ding{55}&47.9\\
24 & 0.70 & \ding{55}&48.2\\
8 & 0.23 & \ding{51}&\best{48.9}\\

\bottomrule
\end{tabular}
}
 
}
\hfill
\mpage{0.28}{
\resizebox{0.9\linewidth}{!}{
\begin{tabular}{lc}
\toprule
Frame sampling strategy  &  Acc.\\
\midrule
Random & 48.2 \\
Temporal attention & 48.6 \\
Uniform & \best{48.9} \\
 
\bottomrule
\end{tabular}
}
    }
\hfill
\mpage{0.28}{
\resizebox{0.7\linewidth}{!}{
\begin{tabular}{lc}
\toprule

Input features & Acc. \\
\midrule
$\S_{\text{dense}}$ & 47.6\\
$\S_{\text{sparse}}$ & 48.9\\

\bottomrule
\end{tabular}
}
}

\hfill
\vspace{-5mm}

\label{tab:abl}
\end{table*}


%% file: 6_conclusion.tex
\vspace{-0.5em}

\section{Conclusions}
\label{sec:conclusions}
\vspace{-0.5em}
In this work, we tackle the challenging problem of video class-incremental learning (VCIL). 
%
We propose \ours{}, a VCIL method designed to achieve a \emph{high memory-efficiency-performance trade-off}.
%
\ours{} consists of episodic memory for storing temporally sparse features, semantic memory for storing general knowledge represented by learnable prompts, and a memory retrieval (MR) module that integrates these two memory components. 
The MR module retrieves temporally dense features from the temporally sparse features stored in episodic memory, through cross-attention to semantic prompt stored in semantic memory.
We validate the effectiveness of \ours{} through extensive experiments on the vCLIMB and TCD benchmarks.
We hope this work inspires future research in video class-incremental learning.


%% file: 99_supp_content.tex
In this supplementary material, we provide 
architecture/implementation/metrics/dataset details, and additional experimental results
to complement the main paper. We organize the supplementary material as follows:

\begin{enumerate}
    \item Architecture Details
    \item Complete implementation details
    \item Evaluation metric details
    \item Additional experimental results
    \item Dataset details

\end{enumerate}

\section{Architecture Details}
In this section, we provide architecture details of \ours{}. We employ CLIP~\cite{clip} as our frame-level visual encoder. We use $L=8$ of clip length and both MR module and semantic prompt is task-specific by default.

\subsection{Details of memory retrieval.}
In \tabref{arch}, we provide stage-wise details of memory retrieval module in \ours{} during the training stage.
We omit the task index $k$ for simplicity.
Given an input video clip $\X$ with a length of 8, we first pass it through the frozen visual encoder, resulting in frame-level features $\dense \in \mathbb{R}^{L\times d} $. 
Note that we omit the notation 'frame-level' for simplicity. 
In this case, we set the number of frame features to $l=2$.
Subsequently, to obtain temporally sparse features $\sparse$, we temporally sub-sample the $\dense$ using randomly selected frame indices.
Then, we take semantic prompt, $\P$, and pass both the semantic prompt $\P$ and the temporally sparse frame-level features  $\sparse$ through the MR module.
The multi-head cross-attention ($\mhca$) layer enables the effectively integration of between $\P$ and $\sparse$.
Afterward, we pass the output tensors from the $\mhca$ layer through the feed-forward network ($\ffn$), resulting the retrieved dense features $\retrieved$. 
%
%
Please refer to the Section 3.2 of the main paper for the remaining training process.

\input{table/supp/architecture}
\subsection{Temporal encoder architecture.}
As shown in Figure 3 of the main paper, we employ a temporal encoder, denoted as $f_t(\cdot; \theta_t)$, to obtain a clip-level feature vector.
The temporal encoder consists of one or more Transformer layers, including attention layers and feed-forward network ($\ffn$).
Among the two design choices for the attention layer: i) self-attention and ii) cross-attention, we adopt cross-attention. 
In the \emph{self-attention} architecture, we concatenate $\sparse$ and semantic prompt $\P$ and pass it through self-attention layers.
In the \emph{cross-attention} architecture, we first initialize a learnable feature vector as a query and pass it through cross-attention layers, using $\sparse$ as a key and value. 
In \tabref{abl2}, we show the experimental results analyzing the effect of different temporal encoder architectures.
In the temporal encoder with cross-attention, we initialize a learnable query token and perform the attention operation, where key and value come from input frame-level features.
The temporal encoder enhances the temporal understanding of frame-level features obtained from the frozen visual encoder, addressing their limitations in capturing temporal context.
\subsection{Architecture details of the baselines in the MR architecture ablation study.}
In \figref{abl_arch}, we visualize the architectures of the baselines used in the MR module architectures ablation study, as presented in Table 3 (b) of the main paper.
In \figref{abl_arch} (a), we present the \emph{MLP w/o semantic memory} baseline, which includes three fully-connected layer with GELU activation without semantic memory.
For the fully-connected layer, we flatten $\sparse$, resulting in $\mathbb{R}^{l\cdot d}$.
After passing the $\sparse$ through \emph{MLP}, we reshape the output tensors, $\retrieved \in \mathbb{R}^{L\cdot d}$ to $\mathbb{R}^{L\times d}$.
In \figref{abl_arch} (b), we present the architectures of \emph{Add} and \emph{Multiply} baselines, which serves as naive baselines to integrate semantic prompt, $\P$ and temporally sparse features,  $\sparse$.
To match a clip length between $\P$ and $\sparse$, we conduct nearest neighbor interpolation $\sparse$ along the temporal axis.
In \figref{abl_arch} (c), we show the \emph{Self-attention} baseline, which replaces the cross-attention layer of MR module with self-attention layer.
We append the $\sparse$ to the $\P$ and pass them through a self-attention layer and a feed-forward network. Afterward, we detach the $\sparse$ from the output tensors to obtain the $\retrieved$.

\section{Implementation details}
\label{sec:detail}
\subsection{Training.}
We use CLIP ViT-B/16 ~\cite{clip} as the visual encoder and keep it frozen. 
The temporal encoder consists of a 3-layer cross-attention architecture, while the memory retrieval module utilizes a 1-layer cross-attention architecture.
For SSV2~\cite{ssv2}, we attach and learn a lightweight adapter~\cite{yang2023aim} during the base task learning stage and then freeze it in the subsequent incremental learning stages.
{We conduct our experiments with a batch size of 24 for all datasets except UCF-101~\cite{soomro2012ucf101}, where the batch size is set to 10. The learning rate is set to 0.001, following a cosine scheduling strategy.}
{For the current task training stage, we train for 50 epochs for all datasets except UCF-101, where we train for 30 epochs. During the rehearsal stage, we train for 30 epochs across all datasets.}
We conduct the experiments with 24 NVIDA GeForce RTX 3090 GPUs for all datasets except UCF-101 and ActivityNet~\cite{caba2015activitynet}, where we use 8 GPUs.
We implement \ours{} using PyTorch and build upon the code of VideoMAE~\cite{tong2022videomae}.
We follow the prior works~\cite{gao2023unified,zhang2023slcaslow} in adopting the local cross-entropy loss, where we only compute the loss between current task logits and ground truth labels and apply logit masking for the classes belong to the other tasks.
We set both the $\alpha$ and $\beta$ to 1.0 for all experiments.
\input{figure/supple/abl_architecture}
\subsection{Memory update.}
\input{figure/supple/memory_update}
After each training stage, given an input clip length of $L$, we store $N_s$ per class feature vectors with a clip length of $l$ ($l \ll L$) in the episodic memory and a semantic prompt with a clip length of $L$ in the semantic memory as shown in \figref{memory_update}.
%
In Tables 1-3 of the main paper, we use the optimal combination of $(N_s \times l)$ for each dataset on two benchmarks~\cite{villa2022vclimb,tcd}. 
In the TCD benchmark, we select $(N_s \times l)$ combinations as follows: UCF-101 ($10\times1$), HMDB51 ($10\times 1$) and SSV2 ($4\times4$).
In the vCLIMB benchmark, we select $(N_s \times l)$ combinations as follows: ActivityNet ($32\times2$), Kinetics-400 ($32\times 2$) and UCF-101 ($16\times2$).
In \tabref{ns_l_datasets}, we show the experimental results exploring how to determine the optimal ($N_s \times l$) configuration.

\section{Evaluation Metric.}
\paragraph{Performance}
%
%
We evaluate the performance using two metrics: Average Accuracy (AA) and Average Incremental Accuracy (AIA). 
AA on $k$-th task, $AA_k$ is the average classification accuracy of the model evaluated up to the $k$-th task.
%
Following vCLIMB~\cite{villa2022vclimb}, we evaluate UCF-101, ActivityNet, and Kinetics-400 using the final average accuracy, $AA_K$, where $K$ denotes the total number of tasks.
%
%
Following TCD~\cite{tcd}, we evaluate UCF-101 and SSV2 using the Average Incremental Accuracy (AIA), defined as $\text{AIA} = \frac{1}{K} \sum_{k=1}^{K} AA_k$, where $K$ denotes the total number of tasks.
%
\paragraph{Memory usage.}
For details on the memory usage of \ours{}, please refer to Section 4.2 of the main paper.
%
We compute the memory usage for storing RGB frames using the formula  $N_s \times L \times 224 \times 224 \times 3 $
where $N_s$ is the number of samples per class and $L$ is the temporal length. We assume that a single frame has a resolution of $ 224 \times 224 $ pixels with 3 color channels.
%
For centroid feature vectors~\cite{stprompt}, we calculate the memory usage as $N_c \times d_c \times 4$, where $N_c$ is the number of centroids, and $d_c$ is the centroid dimension.
%
For the STSP~\cite{chengstsp}, raw examples or features are not stored in episodic memory. Instead, it stores a covariance matrix.  
Thus, we assume that the input consists of frames with a resolution of $224 \times 224$ pixels and then compute the memory usage of the covariance matrix accordingly.
Since there is no publicly available code of ST-Prompt~\cite{stprompt} and STSP~\cite{chengstsp}, we calculate the memory usage based on the configuration described in the paper.~\footnote{
ST-Prompt reports the size of the prompts but not the size of the centroids. 
Additionally, they do not take data types into account.
}
\input{table/supp/sem_memory_usage}
\paragraph{Memory consumption by MR module.}
%
Since we consider only raw RGB frames and feature vectors extracted from video samples when calculating memory usage, we also report the additional memory overhead introduced by MR module for completeness.
%
Each MR module contains approximately 7M parameters, occupying around 27\,MiB in FP32 precision. 
In the \ours{}$_\text{task}$ configuration, where a separate MR module is maintained per task, this results in a total of $\sim$27\,$\times$\,\#Tasks\,MiB memory consumption.

As shown in the Table~\ref{tab:memory}, we report the memory usage of each method (Mem) along with the additional memory required by our MR modules (MR) in vCLIMB UCF-101, 10 tasks setting. 
For a fair comparison, we also include the result that estimates PIVOT's memory usage assuming it stores raw video frames in JPEG-compressed format.
\ours{} is more memory-efficient than storing raw RGB data—with JPEG-compression— (7.5GiB \vs 279.7MiB).

For scenarios where the MR module memory may be a concern, we also provide results of a variant using a single global MR module across tasks (\ours{}$_{global}$).
As shown in Table below and in Table 3  (a) of the main paper, \ours{}$_{global}$ achieves favorable performance (94.8\%) with significantly lower memory usage (27MiB).
%
Optimizer state occupies $\sim$54MiB during the incremental training stage.
%
However, since we freeze the MR module during the rehearsal stage, it does not incur any optimizer memory overhead.
\section{Additional experimental results.}


\subsection{Additional analysis.}
\label{sec:additional_analysis}

\paragraph{Visualization of memory retrieval.}
\input{figure/supple/tsne_wo_semantic}
\input{figure/supple/gpu_memory}
\input{figure/supple/tsne_w_semantic}

We provide a T-SNE visualization for Task 3 of SSV2 ($10\times9$) in \figref{tsne}. 
%
%
Additionally, for a fair comparison, we use the validation set for this analysis.
We compare the $L_2$ distances from the original temporally dense feature vectors ($\dense$) to (i) the temporally sparse feature vectors ($\sparse$) and (ii) the retrieved feature vectors ($\retrieved$). 
As shown in \figref{tsne}, the distance between $\dense$ and $\retrieved$ is significantly smaller than the distance between $\dense$ and $\S_{\text{sparse}}$ (19.8 \vs 8.7).
This result demonstrates that the MR module effectively retrieves temporally dense feature vectors from temporally sparse feature vectors.


\vspace{-1em}
\paragraph{GPU memory usage of MR module.}

The MR module consists of cross attention and MLP components. Each MR module occupies approximately 27MiB of GPU memory when using float32 precision. Notably, we train the k-th MR module during the $k$-th task training stage and then freeze it for subsequent use. We plot the GPU memory usage of \ours{} as the number of task increases as shown in \figref{gpu}. The memory usage remains approximately 15GB even when the number of task reaches 500, which is a reasonably high number in continual learning. This demonstrates the memory-efficiency of \ours{}, even in scenarios with a large number of total tasks.
Additionally, if GPU memory is somehow very limited, 
we can use a global MR module as a memory-efficient alternative while still achieving comparable performance, as shown in Table 4 (a) of the main paper.


\input{figure/supple/ucf_ns_l}
\vspace{-1em}

\paragraph{Visualization of semantic prompt.}
We present the T-SNE~\cite{van2008visualizingtsne} visualization of semantic prompt and task feature vectors to observe what the semantic prompt learns.
As shown in \figref{general}, the semantic prompt before training $\P_\text{before}$ (\textcolor{before}{triangle}) before training is far from the current task feature vectors ($\dense$, $\sparse$, and $\retrieved$ ). 
After training, we observe that the semantic prompt (\textcolor{after}{triangle}) is located at the center of the current task feature vectors. 
This indicates that the semantic prompt captures general knowledge of the current task.

\paragraph{What is the optimal $(N_s \times l)$ configuration for episodic memory?}
In \tabref{ns_l_datasets}, we investigate the optimal configuration of $(N_s \times l)$ for episodic memory, where
$N_s$ and $l$ denote the number of samples stored per class and the number of frame features.
%
Since the UCF-101 is a static-biased dataset~\cite{li2018resound,choi2019can}, reducing the number of frames and storing more samples yields the highest performance. 
On the other hand, we observe that for the SSV2 dataset, \ours{} achieves the highest performance when $N_s$ = 4 and $l$ = 4. The result indicates that \ours{} needs to store more frames than the UCF-101 setting to capture the essential temporal dynamics since the SSV2 is a temporal-biased dataset~\cite{timesformer,kowal2022deeper}.
\input{table/supp/comp_rehearsal_free}
\input{figure/supple/figure_taskwise}
\paragraph{Comparison with rehearsal-free method.}
We compare \ours{} with rehearsal-free baselines in terms of accuracy and final backward forgetting (BWF) on the vCLIMB UCF-101 dataset with 20 tasks. 
As shown in Table ~\ref{tab:rehearsal_free}, \ours{} significantly outperforms CODA-Prompt~\cite{smith2023coda}, a rehearsal-free method, achieving both higher accuracy (95.9\% vs. 74.2\%) and lower forgetting (BWF 2.3 vs. 13.3).
This indicates that \ours{} effectively mitigates forgetting even without relying on rehearsal buffers. 
To further illustrate this, we plot full task-wise accuracy curves—i.e., the diagonal and last row of the full accuracy matrix—in \figref{taskwise}, showing that \ours{} retains substantially higher accuracy as the number of tasks increases.

\subsection{Additional ablation study}
To further validate the effect of each component, we present ablation results on the Something-Something-V2 (10 $\times$ 9 tasks) dataset. 
We report the Top-1 average incremental accuracy (\%). 
We use CLIP~\cite{clip} as a backbone encoder and set $N_s=4$, and $l=4$ for episodic memory. 

\input{table/supp/ablation1}
\paragraph{Ablation study on the impact of $\alpha$ and $\beta$.}
\ours{} calculates the total loss ($L_{\text{total}}$) by applying a weighted sum of the static matching loss ($L_{\text{SM}}$) and temporal matching loss ($L_{\text{TM}}$) using ($\alpha$) and $\beta$ as described in Eq. (10) of the main paper. 
We show the ablation results on ($\alpha$) and $\beta$ in the \tabref{abl1}. 
We achieve the highest performance when we set the value of both hyperparameters to 1.

\input{table/supp/ablation2}
\input{table/supp/ablation4}

\paragraph{Effect of temporal encoder architecture.}
In \tabref{abl2}, using cross-attention in the temporal encoder, $f_t(\cdot; \theta_t)$, shows favorable performance compared to using self-attention.


\paragraph{Effect of using MR module during inference stage.}
For each task, we have a dedicated MR module and dedicated semantic prompt. 
Since we do not have task ID information during inference in the class-incremental learning setting, for each test sample, we have to estimate the task ID information to select MR module and semantic prompt.
Additionally, we provide experimental results by using a global MR module and global semantic prompts so that we do not need to estimate task ID information during testing. 
As shown in \tabref{abl4}, we observe that using the MR module and semantic prompt during inference has little impact on performance (47.4 \vs 47.8). 
Therefore, to improve computation efficiency, we omit the MR module and semantic prompt during testing.

\section{Dataset Details}
In this section, we provide a detailed description of the datasets.
\paragraph{vCLIMB Benchmark}
The vCLIMB benchmark~\cite{villa2022vclimb} includes UCF-101~\cite{soomro2012ucf101}, AcvitiyNet~\cite{caba2015activitynet}, and kinetics-400~\cite{kinetics}.
The vCLIMB benchmark provides two experimental settings: a 10 tasks and a 20 tasks configuration. In the 10 tasks setting, the entire set of action classes is partitioned into 10 sequential tasks, whereas in the 20 tasks setting, the classes are split into 20 sequential tasks.

\paragraph{TCD Benchmark}
The TCD benchmark~\cite{tcd} includes UCF-101~\cite{soomro2012ucf101}, HMDB51~\cite{hmdb}, and SSV2~\cite{ssv2}.
The UCF-101 in TCD consists of a base task learning 51 classes and provides three settings: $10$ classes × $5$ tasks, $5$ classes × $10$ tasks, and $2$ classes × $25$ tasks.
The HMDB51 in TCD consists of a base task learning 26 classes and provides two settings: $5$ classes × $5$ tasks and $1$ class × $25$ tasks.
The SSV2 in TCD consists of a base task with 84 classes and offers two settings: $10$ classes × $9$ tasks and $5$ classes × $18$ tasks.

%% file: table/supp/architecture.tex
\newlength\savewidth\newcommand\shline{\noalign{\global\savewidth\arrayrulewidth\global\arrayrulewidth1.2pt}\hline\noalign{\global\arrayrulewidth\savewidth}}

\newcommand{\greenline}{\arrayrulecolor{green}\hline\arrayrulecolor{black}}
\newcommand{\yellowline}{\arrayrulecolor{yellow}\hline\arrayrulecolor{black}}
\newcommand{\orangeline}{\arrayrulecolor{orange}\hline\arrayrulecolor{black}}
\newcommand{\redline}{\arrayrulecolor{red}\shline\arrayrulecolor{black}}

\newcommand{\tablestyle}[2]{\setlength{\tabcolsep}
{#1}\renewcommand{\arraystretch}{#2}\centering\footnotesize}

\def\x{$\times$}
\definecolor{pastelgreen}{RGB}{119, 221, 119}
\definecolor{scolor}{RGB}{70, 159, 44}
\newcommand{\scolor}[1]{\textcolor{scolor}{#1}}
\definecolor{dcolor}{RGB}{105, 104, 214}
\newcommand{\dcolor}[1]{\textcolor{dcolor}{#1}}
\definecolor{gcolor}{RGB}{232, 160, 20}
\newcommand{\gcolor}[1]{\textcolor{gcolor}{#1}}
\definecolor{tcolor}{RGB}{89,173,196}
\newcommand{\tcolor}[1]{\textcolor{tcolor}{#1}}
\definecolor{bcolor}{RGB}{139,85,28}
\newcommand{\bcolor}[1]{\textcolor{bcolor}{#1}}
\definecolor{redcolor}{RGB}{255,0,0}
\newcommand{\redcolor}[1]{\textcolor{redcolor}{#1}}
    \begin{table}
        \centering
        \caption{\tb{Stage-wise details of the Memory retrieval.} We provide a detailed description of each operation performed from an input clip to frame-level reconstruction. The input for this example is one video clip consisting of 8 frames (skip batch term). In the description, \dcolor{D} and \tcolor{T} represent the embedding dimension and temporal sequence length, respectively. We omit the Layer Normalization and activation functions for simplicity.}
        \tablestyle{1pt}{1.4}
        \resizebox{0.9\linewidth}{!}{
        \begin{tabular}{|c|c|c|}
        
        \hline
        \multicolumn{1}{|c|}{\cellcolor{gray!20}}&
        \multicolumn{2}{c|}{\cellcolor{pastelgreen!70}\textbf{Memory Retrieval Module}}
        \\
        \hline
    
        \cellcolor{gray!20}\textbf{Stage} & 
        \cellcolor{pastelgreen!30}\textbf{\quad\quad Remark \quad\quad} & 
        \cellcolor{pastelgreen!30}\textbf{Output Tensor Shape} 
        \\
        \hline

        Feed-Forward &  Linear Down& \multirow{2}{*}{$\retrieved:\tcolor{8}\times\dcolor{768}$}\\
        Network&\scriptsize{\emph{with ratio} = \gcolor{$0.25$}}&\\ 
        \hline
        Feed-Forward &  Linear Up \& GELU& \multirow{2}{*}{$\P:\tcolor{8}\times \dcolor{3072}$}\\
        Network&\scriptsize{\emph{with ratio} = \gcolor{$4.0$}}&\\ 
        \hline
        \multirow{2}{*}{Cross-Attention} &  $\text{MHCA}(\P,\sparse)$& \multirow{2}{*}{$\P:\tcolor{8}\times\dcolor{768}$}\\
        &\scriptsize{window shape:} \footnotesize{\emph{\tcolor{time}}} &\\
        \hline


        \multirow{2}{*}{Input of MR module} &  Taking $\P$& $\P:\tcolor{8}\times\dcolor{768}$\\
        &\scriptsize{from \emph{semantic memory}}&$\sparse:\tcolor{2}\times\dcolor{768}$\\
        \hline

        Temporal  & Random selection &\multirow{2}{*}{$\sparse:\tcolor{2}\times\dcolor{768}$}\\
        sub-sampling&from $\dense$ &\\
        \hline
        
        Extracting & \multirow{2}{*}{Frozen $f_s(\cdot)$} &\multirow{2}{*}{$\dense:\tcolor{8}\times\dcolor{768} $}\\
        frame-level features && \\
        \hline 
        

          Input video clip & - &$\X:\tcolor{8}\times 224 \times 224 \times 3 $ \\
        \hline 


        \end{tabular}}

        \label{tab:arch}
    \end{table}

%% file: figure/supple/abl_architecture.tex
\begin{figure}[h]
    \centering
    \includegraphics[width=\linewidth]{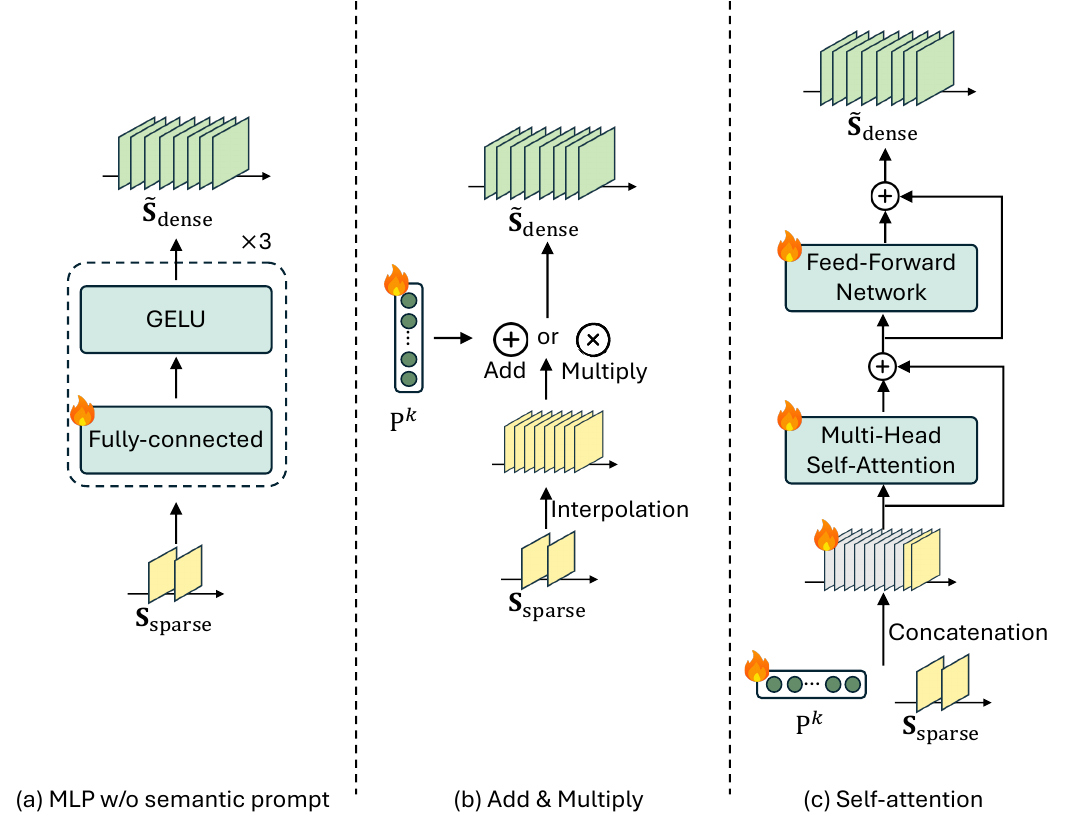}
    \caption{\textbf{Architecture visualization of the different memory retrieval methods.}
    We visualize the various MR architectures presented in Table 3 (b) of the main paper. }
    \label{fig:abl_arch}
\end{figure}

%% file: figure/supple/memory_update.tex
\begin{figure}[h]
    \centering
    \includegraphics[width=0.75\linewidth]{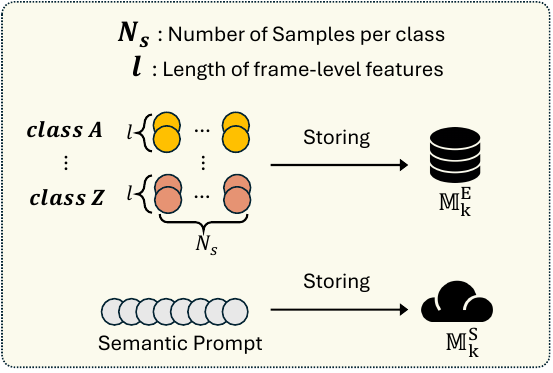}
    \caption{\textbf{Memory Update.}
    After training $k$-th task, we update episodic memory and semantic memory.
    For episodic memory, we select $N_s$ samples per class and subsample $l$ frame-level features from each sample among the training samples of the current task.
    For semantic memory, we store the semantic prompt learned from the current task.
    }
    \label{fig:memory_update}
\end{figure}

%% file: table/supp/sem_memory_usage.tex
\vspace{-0.5em}
\begin{table}[h]
\centering 
\caption{\tb{Memory usage of VCIL methods, including additional overhead from MR modules in \ours{}.}}
\resizebox{\linewidth}{!}{
\begin{tabular}{l rrrr}
\toprule
             Method & Params & Mem.   & MR module & Acc. \\
             \midrule
ST-Prompts & 151M   & 80.5MiB & -                 & 85.1 \\
PIVOT (RAW frame)  & 161M & 26.4GiB & -                 & 93.4 \\
PIVOT (JPEG compressed) & 161M & 7.5GiB & -                 & 93.4 \\
\midrule
\ours{}$_{task}$ &162M  & 9.7MiB   & 270MiB               & \best{95.8} \\
\ours{}$_{global}$ &99M& 9.7MiB   & 27MiB                & 94.8  \\
\bottomrule
\end{tabular}
}
\vspace{-0.5em}

\label{tab:memory}
\end{table}

%% file: figure/supple/tsne_wo_semantic.tex
\begin{figure}
    \centering
    \includegraphics[width=\linewidth]{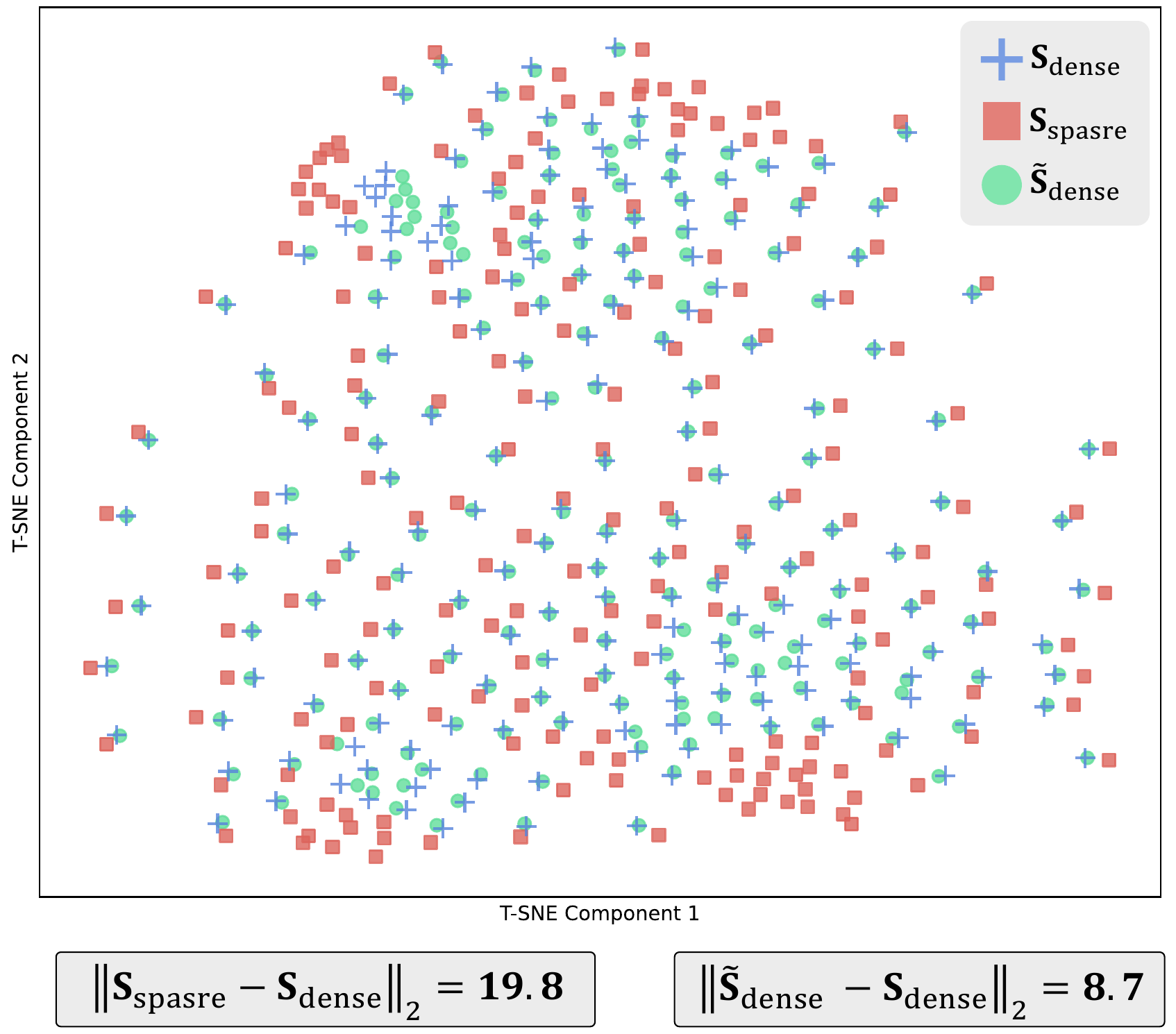}
    \captionsetup{justification=justified, singlelinecheck=false}
    \caption{\textbf{T-SNE~\cite{van2008visualizingtsne} visualization of memory retrieval of the MR module on third task in SSV2 ($10\times9$) tasks.}}
    \label{fig:tsne}
\end{figure}

%% file: figure/supple/gpu_memory.tex
\begin{figure}[h]
    \centering
    \includegraphics[width=\linewidth]{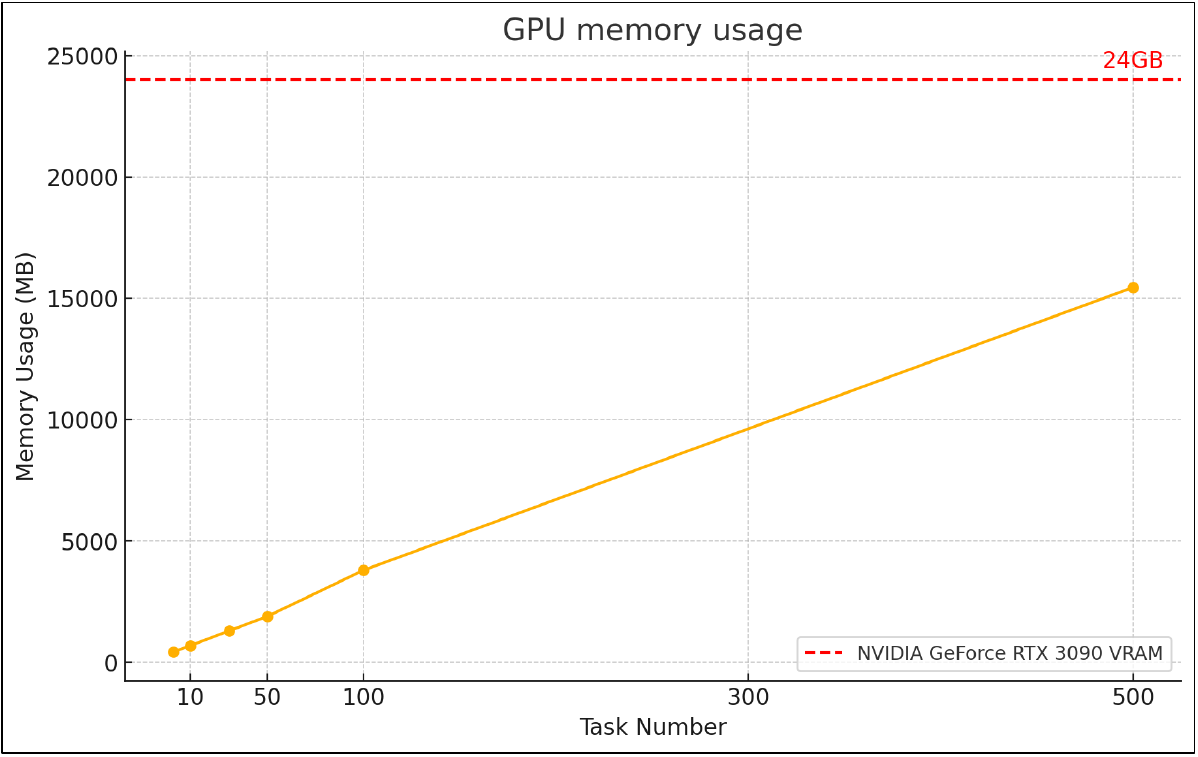}
    \caption{\textbf{GPU memory usage of \ours{} as the number of tasks increases during training on the SSV2 dataset.}
    We plot the GPU memory usage required for training as the number of tasks increases up to 500 on a single RTX 3090 GPU. The GPU memory capacity of the RTX 3090 is represented by a red dashed line. Notably, \ours{} utilizes only approximately 63\% of the available GPU memory even when the number of tasks reaches 500, showcasing its remarkable memory efficiency.
    }
    \label{fig:gpu}
\end{figure}

%% file: figure/supple/tsne_w_semantic.tex
\begin{figure*}[t]
    \centering
    \includegraphics[width=0.8\linewidth]{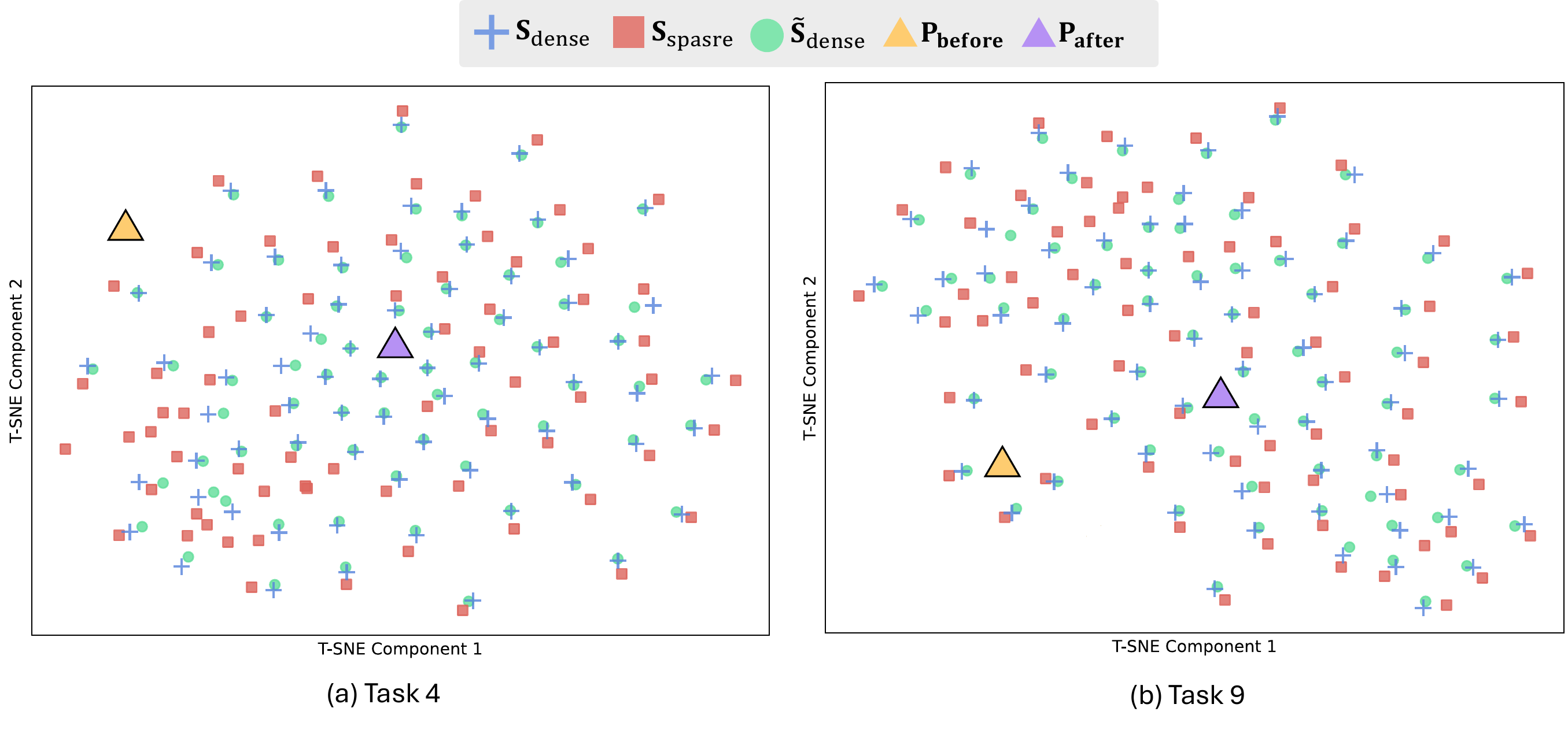}
    \caption{\textbf{T-SNE of semantic prompts and task feature vectors in SSV2 ($10\times9$) tasks.}
We present the T-SNE~\cite{van2008visualizingtsne} visualization of semantic prompt and task feature vectors to observe what the semantic prompt learns.
(a) The semantic prompt before training $\P_\text{before}$ (\textcolor{before}{triangle})  is far from the current data feature vectors ($\dense$,$\sparse$, and $\retrieved$ ). 
After training, we observe that the semantic prompt $\P_\text{after}$ (\textcolor{after}{triangle}) is located at the center of the current data feature vectors (\textcolor{before}{triangle} $\rightarrow$ \textcolor{after}{triangle}). 
(b) Similar to (a), in the ninth task, the semantic prompt also moves toward the center of the task feature vectors after training.
This indicates that the semantic prompt captures general knowledge of the current task.
    }
    \label{fig:general}

\end{figure*}


%% file: figure/supple/ucf_ns_l.tex
\begin{table*}[t]
\centering
\caption{
\textbf{Performance comparison with different memory configurations on UCF-101 and SSV2 datasets of TCD benchmark.}
We report the Top-1 average incremental accuracy (\%) and the total memory usage (MB) for different ($N_s \times l$) configurations. Here, $N_s$ and $l$ denote the number of samples stored per class and the number of frame features. The \textbf{best} results are highlighted.
}
\vspace{-3mm}

\begin{minipage}{0.54\linewidth}
\centering 
\resizebox{\linewidth}{!}{ 
\begin{tabular}{c c c rrrrr}
\toprule
\makecell[l]{\multirowcell{6}[-0.2ex]{Backbone}} &
\makecell[l]{\multirowcell{6}[-0.2ex]{($N_s\times l$)}} & 
\multicolumn{5}{c}{UCF-101} \\ 
\cmidrule{3-8} 
 & & \multicolumn{2}{c}{10 $\times$ 5 Tasks} &\multicolumn{2}{c}{5 $\times$ 10 Tasks}&\multicolumn{2}{c}{2 $\times$ 25 Tasks} \\
\cmidrule(lr){3-4}
\cmidrule(lr){5-6}
\cmidrule(lr){7-8}
&& \shortstack[r]{Memory \\ Usage} & 
 \multirow{-2}{*}{Acc.} &
 \shortstack[r]{Memory \\ Usage} & 
 \multirow{-2}{*}{Acc. }
 &
 \shortstack[r]{Memory \\ Usage} & 
 \multirow{-2}{*}{Acc. }
 \\
\midrule
\multirow{3}{*}{\makecell{ImageNet  \\ pre-trained\\ ViT-B/16}}&($10\times 1$)& \multirow{3}{*}{3.1M}	&\textbf{92.8}	 & \multirow{3}{*}{3.2M}&\textbf{90.8}&\multirow{3}{*}{3.6M}	&\textbf{90.1} \\
&($5\times 2$)& 	&{92.1}	 & &{90.2}& &{89.1} \\
&($2\times 5$)& 	&{90.8}	 & &{88.2}&&{84.1} \\
\midrule
\multirow{3}{*}{\makecell{CLIP  \\ pre-trained\\ ViT-B/16}}&($10\times 1$)& \multirow{3}{*}{3.1M}	&{95.1}&\multirow{3}{*}{3.2M}&\textbf{93.9}& \multirow{3}{*}{3.6M}	&\textbf{93.3} \\
&($5\times 2$)& 	&\textbf{95.3}& &{93.2}& 	&{92.5} \\
&($2\times 5$)& 	&{93.0}& &{91.9}& 	&{89.9} \\
\bottomrule
\end{tabular}
}
\end{minipage}
\hfill
\begin{minipage}{0.42\linewidth}
\centering 
\resizebox{\linewidth}{!}{ 
\begin{tabular}{c c c rrrrr}
\toprule
\makecell[l]{\multirowcell{6}[-0.2ex]{Backbone}} &
\makecell[l]{\multirowcell{6}[-0.2ex]{($N_s\times l$)}} & 
\multicolumn{5}{c}{SSV2} \\ 
\cmidrule{3-8} 
 & & \multicolumn{2}{c}{10 $\times$ 9 Tasks} &\multicolumn{2}{c}{5 $\times$ 18 Tasks} \\
\cmidrule(lr){3-4}
\cmidrule(lr){5-6}
&& \shortstack[r]{Memory \\ Usage} & 
 \multirow{-2}{*}{Acc.} &
 \shortstack[r]{Memory \\ Usage} & 
 \multirow{-2}{*}{Acc. }
 \\
\midrule
\multirow{3}{*}{\makecell{ImageNet  \\ pre-trained\\ ViT-B/16}}&($8\times 2$)& \multirow{3}{*}{8.4M}	&{42.3}	 & \multirow{3}{*}{8.6M}&{40.8} \\
&($4\times 4$)& 	&\textbf{44.9}	 & &\textbf{42.7} \\
&($2\times 8$)& 	&{40.2}	 & &{38.5} \\
\midrule
\multirow{3}{*}{\makecell{CLIP  \\ pre-trained\\ ViT-B/16}}&($8\times 2$)& \multirow{3}{*}{8.4M}	&{47.1}&\multirow{3}{*}{8.6M}&{45.8} \\
&($4\times 4$)& 	&\textbf{48.9}& &\textbf{47.5} \\
&($2\times 8$)& 	&{45.5}& &{44.6} \\
\bottomrule
\end{tabular}
}
\end{minipage}
\label{tab:ns_l_datasets}
\vspace{-3mm}
\end{table*}

%% file: table/supp/comp_rehearsal_free.tex
\vspace{-0.5em}
\begin{table}[h]
\centering 
\caption{\tb{\ours{} outperforms rehearsal-free methods in both accuracy and BWF on vCLIMB UCF-101 (20 tasks).}}
\resizebox{0.7\linewidth}{!}{
\begin{tabular}{l c rr}
\toprule
Method      & Rehearsal? & Acc. $\uparrow$   & BWF $\downarrow$    \\
\midrule
CODA-Prompt [\green{46}] & $\times$          & 74.2   & 13.3    \\
\midrule
PIVOT [\green{55}]      & \checkmark          & 93.1  & 3.9     \\
\ours{}   & \checkmark          & \best{95.9}   & \best{2.3}   \\
\bottomrule
\end{tabular}
}
\label{tab:rehearsal_free}
\end{table}
\vspace{-0.5em}

%% file: figure/supple/figure_taskwise.tex
\begin{figure*}[t]
  \centering
  \includegraphics[width=\linewidth]{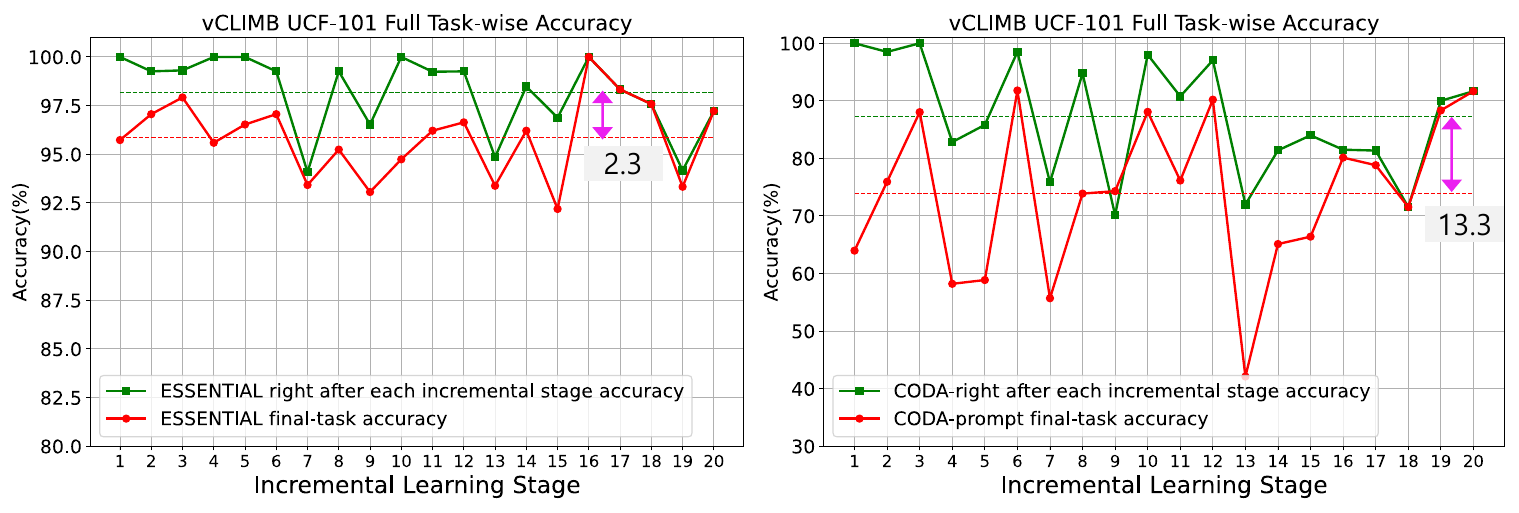}
  \vspace{-2em}
  \caption{\tb{Task-wise accuracy curves on vCLIMB UCF-101, 20 tasks setting.}}
  \label{fig:taskwise}
\end{figure*}

%% file: table/supp/ablation1.tex
\begin{table}[h]
\centering
\caption{\textbf{Ablation study on the impact of $\alpha$ and $\beta$.} 
}

\mpage{1.0}{
\resizebox{0.8\linewidth}{!}{
\begin{tabular}{cc c}
\toprule
\multicolumn{2}{c}{Loss} & \multirow{2}{*}{Acc.} \\
\cmidrule{0-1}
Static matching ($\alpha$) & Temporal matching ($\beta$) &  \\
\midrule
0 & 0 & 46.3 \\
1 & 0 & 46.9 \\
0 & 1 & 47.1 \\
0.5 & 1 & 47.5 \\
1 & 0.5 & 48.1 \\
1 & 1 & \textbf{48.9} \\
\bottomrule
\end{tabular}
}
}
\label{tab:abl1}
\end{table}


%% file: table/supp/ablation2.tex
\begin{table}[t]
\centering
\caption{\textbf{Ablation study on temporal encoder architecture.} 
}

\mpage{1.0}{
\resizebox{0.45\linewidth}{!}{
\begin{tabular}{lcc}
\toprule

Attention strategy  & Acc. \\
\midrule
Self-attention & 48.1 \\
Cross-attention & \best{48.9}  \\
\bottomrule
\end{tabular}
}
}\hfill

\label{tab:abl2}
\end{table}


%% file: table/supp/ablation4.tex
\begin{table}[t]
\centering
\caption{\textbf{Effect of using MR module during inference stage.} 
}

\mpage{1.0}{
\resizebox{0.85\linewidth}{!}{
\begin{tabular}{cccc}
\toprule
MR module&Sem. prompt & Inference w/ MR module & Acc.\\ 
\midrule
Global& Global& $\times$& 47.4\\ 
Global& Global& \checkmark & 47.8\\ 
Task& Task& $\times$ & \textbf{48.9}\\ 

\bottomrule
\end{tabular}
}
}

\label{tab:abl4}
\end{table}


%% file: main.bbl
\begin{thebibliography}{10}\itemsep=-1pt

\bibitem{alssum2023just}
Lama Alssum, Juan~Leon Alcazar, Merey Ramazanova, Chen Zhao, and Bernard Ghanem.
\newblock Just a glimpse: Rethinking temporal information for video continual learning.
\newblock In {\em CVPR Workshop}, 2023.

\bibitem{arnab2021vivit}
Anurag Arnab, Mostafa Dehghani, Georg Heigold, Chen Sun, Mario Lu{\v{c}}i{\'c}, and Cordelia Schmid.
\newblock Vivit: A video vision transformer.
\newblock In {\em ICCV}, 2021.

\bibitem{layernorm}
Jimmy~Lei Ba.
\newblock Layer normalization.
\newblock {\em arXiv preprint arXiv:1607.06450}, 2016.

\bibitem{timesformer}
Gedas Bertasius, Heng Wang, and Lorenzo Torresani.
\newblock Is space-time attention all you need for video understanding?
\newblock In {\em ICML}, 2021.

\bibitem{buzzega2020dark}
Pietro Buzzega, Matteo Boschini, Angelo Porrello, Davide Abati, and Simone Calderara.
\newblock Dark experience for general continual learning: a strong, simple baseline.
\newblock In {\em NeurIPS}, 2020.

\bibitem{caba2015activitynet}
Fabian Caba~Heilbron, Victor Escorcia, Bernard Ghanem, and Juan Carlos~Niebles.
\newblock Activitynet: A large-scale video benchmark for human activity understanding.
\newblock In {\em CVPR}, 2015.

\bibitem{i3d}
Joao Carreira and Andrew Zisserman.
\newblock Quo vadis, action recognition? a new model and the kinetics dataset.
\newblock In {\em CVPR}, 2017.

\bibitem{chengstsp}
Hao Cheng, Siyuan Yang, Chong Wang, Joey~Tianyi Zhou, Alex~C Kot, and Bihan Wen.
\newblock Stsp: Spatial-temporal subspace projection for video class-incremental learning.
\newblock In {\em ECCV}, 2024.

\bibitem{choi2019can}
Jinwoo Choi, Chen Gao, Joseph~CE Messou, and Jia-Bin Huang.
\newblock Why can't i dance in the mall? learning to mitigate scene bias in action recognition.
\newblock In {\em NeurIPS}, 2019.

\bibitem{donahue2016longterm}
Jeff Donahue, Lisa~Anne Hendricks, Marcus Rohrbach, Subhashini Venugopalan, Sergio Guadarrama, Kate Saenko, and Trevor Darrell.
\newblock {Long-term Recurrent Convolutional Networks for Visual Recognition and Description}.
\newblock In {\em CVPR}, 2015.

\bibitem{vit}
Alexey Dosovitskiy, Lucas Beyer, Alexander Kolesnikov, Dirk Weissenborn, Xiaohua Zhai, Thomas Unterthiner, Mostafa Dehghani, Matthias Minderer, Georg Heigold, Sylvain Gelly, et~al.
\newblock An image is worth 16x16 words: Transformers for image recognition at scale.
\newblock In {\em ICLR}, 2021.

\bibitem{eichenbaum2001hippocampus}
Howard Eichenbaum.
\newblock The hippocampus and declarative memory: cognitive mechanisms and neural codes.
\newblock {\em Behavioural brain research}, 127(1-2):199--207, 2001.

\bibitem{fan2021multiscale}
Haoqi Fan, Bo Xiong, Karttikeya Mangalam, Yanghao Li, Zhicheng Yan, Jitendra Malik, and Christoph Feichtenhofer.
\newblock Multiscale vision transformers.
\newblock In {\em ICCV}, 2021.

\bibitem{feichtenhofer2020x3d}
Christoph Feichtenhofer.
\newblock X3d: Expanding architectures for efficient video recognition.
\newblock In {\em CVPR}, 2020.

\bibitem{feichtenhofer2019slowfast}
Christoph Feichtenhofer, Haoqi Fan, Jitendra Malik, and Kaiming He.
\newblock Slowfast networks for video recognition.
\newblock In {\em ICCV}, 2019.

\bibitem{gao2023unified}
Qiankun Gao, Chen Zhao, Yifan Sun, Teng Xi, Gang Zhang, Bernard Ghanem, and Jian Zhang.
\newblock A unified continual learning framework with general parameter-efficient tuning.
\newblock In {\em ICCV}, 2023.

\bibitem{girdhar2022omnivore}
Rohit Girdhar, Mannat Singh, Nikhila Ravi, Laurens van~der Maaten, Armand Joulin, and Ishan Misra.
\newblock {Omnivore: A Single Model for Many Visual Modalities}.
\newblock In {\em CVPR}, 2022.

\bibitem{goodfellow2013empirical}
Ian~J Goodfellow, Mehdi Mirza, Da Xiao, Aaron Courville, and Yoshua Bengio.
\newblock An empirical investigation of catastrophic forgetting in gradient-based neural networks.
\newblock In {\em ICLR}, 2014.

\bibitem{ssv2}
Raghav Goyal, Samira Ebrahimi~Kahou, Vincent Michalski, Joanna Materzynska, Susanne Westphal, Heuna Kim, Valentin Haenel, Ingo Fruend, Peter Yianilos, Moritz Mueller-Freitag, et~al.
\newblock The" something something" video database for learning and evaluating visual common sense.
\newblock In {\em ICCV}, 2017.

\bibitem{hou2019learningUCIR}
Saihui Hou, Xinyu Pan, Chen~Change Loy, Zilei Wang, and Dahua Lin.
\newblock Learning a unified classifier incrementally via rebalancing.
\newblock In {\em CVPR}, 2019.

\bibitem{ji20123d}
Shuiwang Ji, Wei Xu, Ming Yang, and Kai Yu.
\newblock {3D convolutional neural networks for human action recognition}.
\newblock {\em TPAMI}, 35(1):221--231, 2013.

\bibitem{karpathy20142dcnn}
Andrej Karpathy, George Toderici, Sanketh Shetty, Thomas Leung, Rahul Sukthankar, and Li Fei-Fei.
\newblock {Large-scale video classification with convolutional neural networks}.
\newblock In {\em CVPR}, 2014.

\bibitem{kinetics}
Will Kay, Joao Carreira, Karen Simonyan, Brian Zhang, Chloe Hillier, Sudheendra Vijayanarasimhan, Fabio Viola, Tim Green, Trevor Back, Paul Natsev, et~al.
\newblock The kinetics human action video dataset.
\newblock {\em arXiv preprint arXiv:1705.06950}, 2017.

\bibitem{kowal2022deeper}
Matthew Kowal, Mennatullah Siam, Md~Amirul Islam, Neil~DB Bruce, Richard~P Wildes, and Konstantinos~G Derpanis.
\newblock A deeper dive into what deep spatiotemporal networks encode: Quantifying static vs. dynamic information.
\newblock In {\em CVPR}, 2022.

\bibitem{hmdb}
Hildegard Kuehne, Hueihan Jhuang, Est{\'\i}baliz Garrote, Tomaso Poggio, and Thomas Serre.
\newblock Hmdb: a large video database for human motion recognition.
\newblock In {\em ICCV}, 2011.

\bibitem{lee2024cast}
Dongho Lee, Jongseo Lee, and Jinwoo Choi.
\newblock Cast: Cross-attention in space and time for video action recognition.
\newblock In {\em NeurIPS}, 2023.

\bibitem{li2018resound}
Yingwei Li, Yi Li, and Nuno Vasconcelos.
\newblock Resound: Towards action recognition without representation bias.
\newblock In {\em ECCV}, 2018.

\bibitem{liang2024hypercorrelation}
Sen Liang, Kai Zhu, Wei Zhai, Zhiheng Liu, and Yang Cao.
\newblock Hypercorrelation evolution for video class-incremental learning.
\newblock In {\em AAAI}, 2024.

\bibitem{lin2019tsm}
Ji Lin, Chuang Gan, and Song Han.
\newblock {TSM: Temporal Shift Module for Efficient Video Understanding}.
\newblock In {\em ICCV}, 2019.

\bibitem{standalone}
Fuchen Long, Zhaofan Qiu, Yingwei Pan, Ting Yao, Jiebo Luo, and Tao Mei.
\newblock Stand-alone inter-frame attention in video models.
\newblock In {\em CVPR}, 2022.

\bibitem{mccloskey1989catastrophic}
Michael McCloskey and Neal~J Cohen.
\newblock Catastrophic interference in connectionist networks: The sequential learning problem.
\newblock In {\em Psychology of learning and motivation}. Elsevier, 1989.

\bibitem{menon2002relating}
Vinod Menon, Jesse~M Boyett-Anderson, Alan~F Schatzberg, and Allan~L Reiss.
\newblock Relating semantic and episodic memory systems.
\newblock {\em Cognitive brain research}, 13(2):261--265, 2002.

\bibitem{ng2015short}
Joe Yue-Hei Ng, Matthew Hausknecht, Sudheendra Vijayanarasimhan, Oriol Vinyals, Rajat Monga, and George Toderici.
\newblock {Beyond Short Snippets: Deep Networks for Video Classification}.
\newblock In {\em CVPR}, 2015.

\bibitem{pan2022st}
Junting Pan, Ziyi Lin, Xiatian Zhu, Jing Shao, and Hongsheng Li.
\newblock St-adapter: Parameter-efficient image-to-video transfer learning.
\newblock In {\em NeurIPS}, 2022.

\bibitem{tcd}
Jaeyoo Park, Minsoo Kang, and Bohyung Han.
\newblock Class-incremental learning for action recognition in videos.
\newblock In {\em ICCV}, 2021.

\bibitem{motionformer}
Mandela Patrick, Dylan Campbell, Yuki Asano, Ishan Misra, Florian Metze, Christoph Feichtenhofer, Andrea Vedaldi, and Joao~F Henriques.
\newblock Keeping your eye on the ball: Trajectory attention in video transformers.
\newblock In {\em NeurIPS}, 2021.

\bibitem{framemaker}
Yixuan Pei, Zhiwu Qing, Jun Cen, Xiang Wang, Shiwei Zhang, Yaxiong Wang, Mingqian Tang, Nong Sang, and Xueming Qian.
\newblock Learning a condensed frame for memory-efficient video class-incremental learning.
\newblock In {\em NeurIPS}, 2022.

\bibitem{stprompt}
Yixuan Pei, Zhiwu Qing, Shiwei Zhang, Xiang Wang, Yingya Zhang, Deli Zhao, and Xueming Qian.
\newblock Space-time prompting for video class-incremental learning.
\newblock In {\em ICCV}, 2023.

\bibitem{prabhu2020gdumb}
Ameya Prabhu, Philip~HS Torr, and Puneet~K Dokania.
\newblock Gdumb: A simple approach that questions our progress in continual learning.
\newblock In {\em ECCV}, 2020.

\bibitem{clip}
Alec Radford, Jong~Wook Kim, Chris Hallacy, Aditya Ramesh, Gabriel Goh, Sandhini Agarwal, Girish Sastry, Amanda Askell, Pamela Mishkin, Jack Clark, et~al.
\newblock Learning transferable visual models from natural language supervision.
\newblock In {\em ICML}, 2021.

\bibitem{rebuffi2017icarl}
Sylvestre-Alvise Rebuffi, Alexander Kolesnikov, Georg Sperl, and Christoph~H Lampert.
\newblock icarl: Incremental classifier and representation learning.
\newblock In {\em CVPR}, 2017.

\bibitem{reda2022film}
Fitsum Reda, Janne Kontkanen, Eric Tabellion, Deqing Sun, Caroline Pantofaru, and Brian Curless.
\newblock Film: Frame interpolation for large motion.
\newblock In {\em ECCV}, 2022.

\bibitem{sevilla2021only}
Laura Sevilla-Lara, Shengxin Zha, Zhicheng Yan, Vedanuj Goswami, Matt Feiszli, and Lorenzo Torresani.
\newblock Only time can tell: Discovering temporal data for temporal modeling.
\newblock In {\em WACV}, 2021.

\bibitem{Simonyan-NIPS-2014}
Karen Simonyan and Andrew Zisserman.
\newblock Two-stream convolutional networks for action recognition in videos.
\newblock In {\em NeurIPS}, 2014.

\bibitem{smith2023coda}
James~Seale Smith, Leonid Karlinsky, Vyshnavi Gutta, Paola Cascante-Bonilla, Donghyun Kim, Assaf Arbelle, Rameswar Panda, Rogerio Feris, and Zsolt Kira.
\newblock Coda-prompt: Continual decomposed attention-based prompting for rehearsal-free continual learning.
\newblock In {\em CVPR}, 2023.

\bibitem{soomro2012ucf101}
Khurram Soomro, Amir~Roshan Zamir, and Mubarak Shah.
\newblock Ucf101: A dataset of 101 human actions classes from videos in the wild.
\newblock {\em arXiv preprint arXiv:1212.0402}, 2012.

\bibitem{tan2024semantically}
Yuwen Tan, Qinhao Zhou, Xiang Xiang, Ke Wang, Yuchuan Wu, and Yongbin Li.
\newblock Semantically-shifted incremental adapter-tuning is a continual vitransformer.
\newblock In {\em CVPR}, 2024.

\bibitem{tong2022videomae}
Zhan Tong, Yibing Song, Jue Wang, and Limin Wang.
\newblock Videomae: Masked autoencoders are data-efficient learners for self-supervised video pre-training.
\newblock In {\em NeurIPS}, 2022.

\bibitem{tran2015c3d}
Du Tran, Lubomir Bourdev, Rob Fergus, Lorenzo Torresani, and Manohar Paluri.
\newblock {Learning spatiotemporal features with 3d convolutional networks}.
\newblock In {\em ICCV}, 2015.

\bibitem{tran2018closer}
Du Tran, Heng Wang, Lorenzo Torresani, Jamie Ray, Yann LeCun, and Manohar Paluri.
\newblock A closer look at spatiotemporal convolutions for action recognition.
\newblock In {\em CVPR}, 2018.

\bibitem{van2008visualizingtsne}
Laurens Van~der Maaten and Geoffrey Hinton.
\newblock Visualizing data using t-sne.
\newblock {\em JMLR}, 9(11):2579--2605, 2008.

\bibitem{villa2023pivot}
Andr{\'e}s Villa, Juan~Le{\'o}n Alc{\'a}zar, Motasem Alfarra, Kumail Alhamoud, Julio Hurtado, Fabian~Caba Heilbron, Alvaro Soto, and Bernard Ghanem.
\newblock Pivot: Prompting for video continual learning.
\newblock In {\em CVPR}, 2023.

\bibitem{villa2022vclimb}
Andr{\'e}s Villa, Kumail Alhamoud, Victor Escorcia, Fabian Caba, Juan~Le{\'o}n Alc{\'a}zar, and Bernard Ghanem.
\newblock vclimb: A novel video class incremental learning benchmark.
\newblock In {\em CVPR}, 2022.

\bibitem{wang2021triple}
Liyuan Wang, Bo Lei, Qian Li, Hang Su, Jun Zhu, and Yi Zhong.
\newblock Triple-memory networks: A brain-inspired method for continual learning.
\newblock {\em TNNLS}, 33(5):1925--1934, 2021.

\bibitem{wang2018non}
Xiaolong Wang, Ross Girshick, Abhinav Gupta, and Kaiming He.
\newblock Non-local neural networks.
\newblock In {\em CVPR}, 2018.

\bibitem{wang2022s}
Yabin Wang, Zhiwu Huang, and Xiaopeng Hong.
\newblock S-prompts learning with pre-trained transformers: An occam’s razor for domain incremental learning.
\newblock In {\em NeurIPS}, 2022.

\bibitem{wang2022dualprompt}
Zifeng Wang, Zizhao Zhang, Sayna Ebrahimi, Ruoxi Sun, Han Zhang, Chen-Yu Lee, Xiaoqi Ren, Guolong Su, Vincent Perot, Jennifer Dy, et~al.
\newblock Dualprompt: Complementary prompting for rehearsal-free continual learning.
\newblock In {\em ECCV}, 2022.

\bibitem{wang2022learningl2p}
Zifeng Wang, Zizhao Zhang, Chen-Yu Lee, Han Zhang, Ruoxi Sun, Xiaoqi Ren, Guolong Su, Vincent Perot, Jennifer Dy, and Tomas Pfister.
\newblock Learning to prompt for continual learning.
\newblock In {\em CVPR}, 2022.

\bibitem{wiggs1998neural}
Cheri~L Wiggs, Jill Weisberg, and Alex Martin.
\newblock Neural correlates of semantic and episodic memory retrieval.
\newblock {\em Neuropsychologia}, 37(1):103--118, 1998.

\bibitem{wu2022memvit}
Chao-Yuan Wu, Yanghao Li, Karttikeya Mangalam, Haoqi Fan, Bo Xiong, Jitendra Malik, and Christoph Feichtenhofer.
\newblock Memvit: Memory-augmented multiscale vision transformer for efficient long-term video recognition.
\newblock In {\em CVPR}, 2022.

\bibitem{wu2019large}
Yue Wu, Yinpeng Chen, Lijuan Wang, Yuancheng Ye, Zicheng Liu, Yandong Guo, and Yun Fu.
\newblock Large scale incremental learning.
\newblock In {\em CVPR}, 2019.

\bibitem{Xie-ECCV-2018}
Saining Xie, Chen Sun, Jonathan Huang, Zhuowen Tu, and Kevin Murphy.
\newblock Rethinking spatiotemporal feature learning for video understanding.
\newblock In {\em ECCV}, 2018.

\bibitem{yan2022multiview}
Shen Yan, Xuehan Xiong, Anurag Arnab, Zhichao Lu, Mi Zhang, Chen Sun, and Cordelia Schmid.
\newblock Multiview transformers for video recognition.
\newblock In {\em CVPR}, 2022.

\bibitem{yang2023aim}
Taojiannan Yang, Yi Zhu, Yusheng Xie, Aston Zhang, Chen Chen, and Mu Li.
\newblock Aim: Adapting image models for efficient video action recognition.
\newblock In {\em ICLR}, 2023.

\bibitem{zhang2023slcaslow}
Gengwei Zhang, Liyuan Wang, Guoliang Kang, Ling Chen, and Yunchao Wei.
\newblock Slca: Slow learner with classifier alignment for continual learning on a pre-trained model.
\newblock In {\em ICCV}, 2023.

\bibitem{zhao2021video}
Hanbin Zhao, Xin Qin, Shihao Su, Yongjian Fu, Zibo Lin, and Xi Li.
\newblock When video classification meets incremental classes.
\newblock In {\em ACM MM}, 2021.

\bibitem{zhou2018trn}
Bolei Zhou, Alex Andonian, Aude Oliva, and Antonio Torralba.
\newblock {Temporal Relational Reasoning in Videos}.
\newblock In {\em ECCV}, 2018.

\end{thebibliography}
